\documentclass[journal]{IEEEtran}
\usepackage{times}
\usepackage{epsfig}
\usepackage{graphicx}
\usepackage{amsmath}
\usepackage{amssymb}
\usepackage[ruled]{algorithm2e}
\usepackage{algpseudocode}
\usepackage{caption}
\usepackage{subfigure}
\usepackage{booktabs}
\usepackage{tabularx}
\usepackage{xcolor}
\usepackage{multirow}
\usepackage{hyperref}
\usepackage{subfigure}

\begin{document}

\title{SenseFi: A Library and Benchmark on Deep-Learning-Empowered WiFi Human Sensing}

\author{Jianfei~Yang,
	Xinyan~Chen,
	Han~Zou,
	Dazhuo~Wang,
	Chris~Xiaoxuan~Lu,\\
	Sumei Sun,~\IEEEmembership{Fellow,~IEEE}
 	and~Lihua~Xie,~\IEEEmembership{Fellow,~IEEE}
 
 \thanks{
	J. Yang, X. Chen, D. Wang, H. Zou and L. Xie are with the School of Electrical and Electronics Engineering, Nanyang Technological University, Singapore (e-mail: \{yang0478,chen1328,dazhuo001,zouh0005,elhxie\}@ntu.edu.sg).
 	
 	C. X. Lu is with the School of Informatics at the University of Edinburgh, United Kingdom (e-mail: xiaoxuan.lu@ed.ac.uk).
 	
 	S. Sun are with the Institute for Infocomm Research (I$^2$R), Agency for Science, Technology and Research (A*STAR), Singapore 138632.(E-mail: sunsm@i2r.a-star.edu.sg).
	
	}
}

\markboth{}%
{Yang \MakeLowercase{\textit{et al.}}: Deep Learning and Its Applications to Ubiquitous WiFi Sensing: A Benchmark and A Survey}


\maketitle

\begin{abstract}
WiFi sensing has been evolving rapidly in recent years. Empowered by propagation models and deep learning methods, many challenging applications are realized such as WiFi-based human activity recognition and gesture recognition. However, in contrast to deep learning for visual recognition and natural language processing, no sufficiently comprehensive public benchmark exists. In this paper, we review the recent progress on deep learning enabled WiFi sensing, and then propose a benchmark, SenseFi, to study the effectiveness of various deep learning models for WiFi sensing. These advanced models are compared in terms of distinct sensing tasks, WiFi platforms, recognition accuracy, model size, computational complexity, feature transferability, and adaptability of unsupervised learning. It is also regarded as a tutorial for deep learning based WiFi sensing, starting from CSI hardware platform to sensing algorithms. The extensive experiments provide us with experiences in deep model design, learning strategy skills and training techniques for real-world applications. To the best of our knowledge, this is the first benchmark with an open-source library for deep learning in WiFi sensing research. The benchmark codes are available at \href{https://github.com/xyanchen/WiFi-CSI-Sensing-Benchmark}{https://github.com/xyanchen/WiFi-CSI-Sensing-Benchmark}.
\end{abstract}


%
\IEEEpeerreviewmaketitle

\section{Introduction}
With the proliferation of mobile internet usage, WiFi access point (AP) has become a ubiquitous infrastructure in smart environments, ranging from commercial buildings to domestic settings. By analysing the patterns of its wireless signal, today's AP has evolved beyond a pure WiFi router, but is also widely used as a type of `sensor device' to enable new services for human sensing. Particularly, recent studies have found that WiFi signals in the form of Channel State Information (CSI)~\cite{halperin2011tool,xie2015precise} are extremely promising for a variety of device-free human sensing tasks, such as occupancy detection~\cite{zou2017non}, activity recognition~\cite{wang2014eyes,zou2018deepsense,yang2018carefi,zou2019wificv}, fall detection~\cite{wang2016rt}, gesture recognition~\cite{yang2019learning,zou2018gesture}, human identification~\cite{zou2018identification,wang2022caution}, and people counting~\cite{zou2018device,FreeCount}.
Unlike the coarse-grained received signal strengths, WiFi CSI records more fine-grained information about how a signal propagates between WiFi devices and how a signal is reflected from the surrounding environment in which humans move around. 
On the other side, as WiFi signals (2.4GHz or 5GHz) lie in the non-visible band of the electromagnetic spectrum, WiFi CSI based human sensing is intrinsically more privacy-friendly than cameras and draws increasing attention from both academia and industry. Motivated by increasing interests needs, a new WiFi standard, 802.11bf~\cite{802.11bf} is designed by the IEEE 802.11bf Task Group (TGbf), and will amend the current WiFi standard both at the Medium Access Control (MAC) and Physical Layer (PHY) to officially include WiFi sensing as part of a regular WiFi service by late 2024.

\begin{figure}[t]
	\centering
	\includegraphics[width=0.45\textwidth]{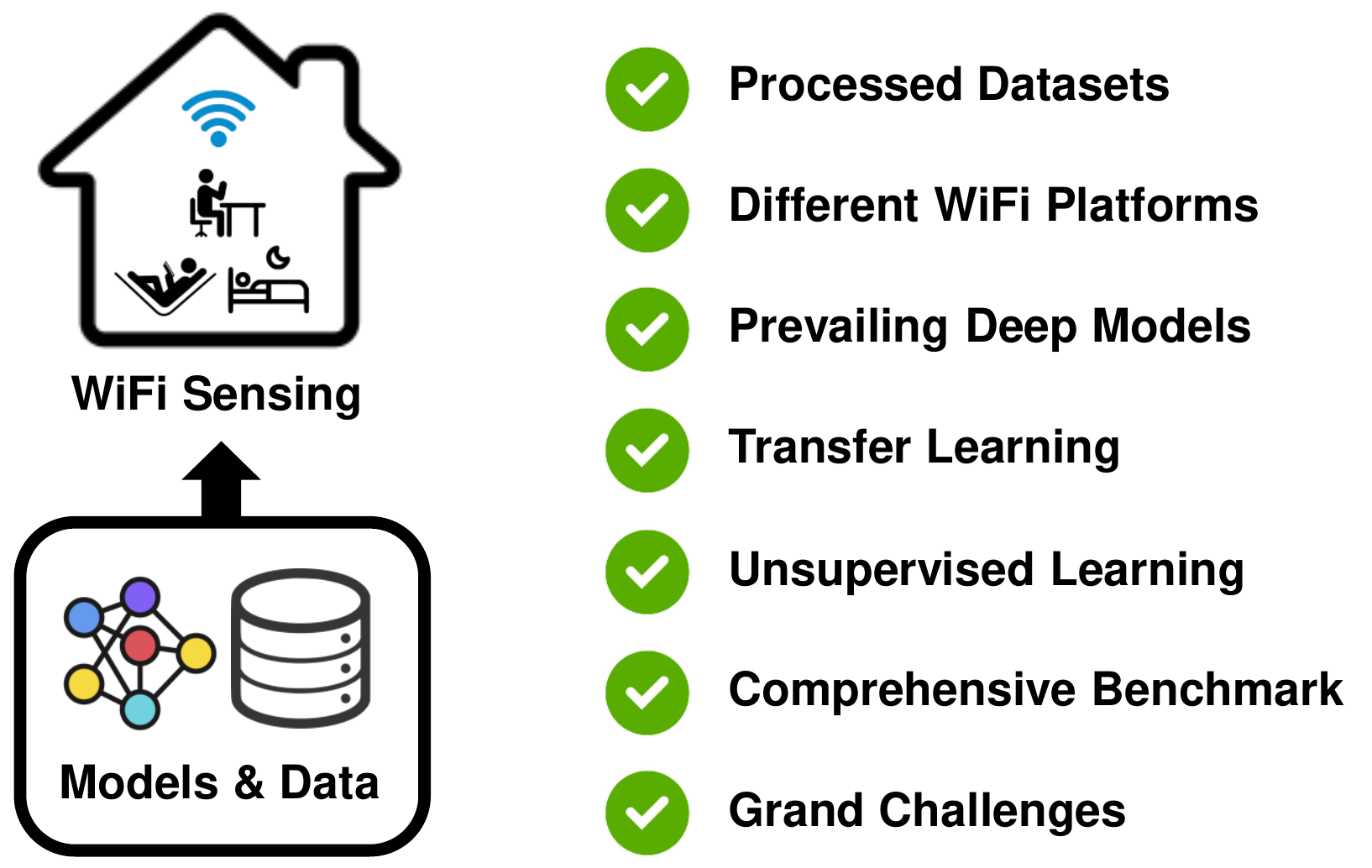}
	\caption{The technical contributions and summary of SenseFi.}\label{fig:principle}
\end{figure}

Existing WiFi sensing methods can be categorized into model-based methods and learning-based methods. Model-based methods rely on physical models that describe the WiFi signals propagation, such as Fresnel Zone~\cite{wu2017device}. Model based methods help us understand the underlying mechanism of WiFi sensing and design sensing methods for periodic or single motions, such as respiration~\cite{wang2016human,8210837,9341474} and falling down~\cite{wang2016rt,9322323,9716074}. Nevertheless, model based methods fall short when it comes to the complicated human activities that consist of a series of different motions.
For example, a human gait comprises the synergistic movements of arms, legs and bodies, the differences of which are hard to depict by physical models. In contrast, by feeding a massive amount of data into machine learning~\cite{yang2018device} or deep learning networks,~\cite{yang2019learning,zou2018deepsense}, learning based achieve remarkable performances in complicated sensing tasks. Various deep neural networks are designed to enable many applications including activity recognition~\cite{zou2017multiple}, gesture recognition~\cite{yang2019learning}, human identification~\cite{zou2018identification,wang2022caution,zhang2020gate}, and people counting~\cite{zou2018device, FreeCount}. Though deep learning models have a strong ability of function approximation, they require tremendous labeled data that is expensive to collect and suffer from the negative effect of distribution shift caused by environmental dynamics~\cite{zou2018robust}. 

Most state-of-the-art deep learning models are developed for computer vision~\cite{voulodimos2018deep}, such as human activity recognition~\cite{shu2020host,zhu2019redundancy}, and natural language processing tasks~\cite{otter2020survey}, such as sentiment classification~\cite{wang2020coarse}. Deep models have demonstrated the capacity of processing high-dimensional and multi-modal data problems. These approaches inspire the deep learning applications in WiFi sensing in terms of data preprocessing, network design, and learning objectives. It is seen that more and more deep models~\cite{bu2021transfersense,zhang2018crosssense} for WiFi sensing come into existence and overcome the aforementioned obstacles that traditional statistical learning methods cannot address. However, current works mainly aim to achieve high accuracy on specific sensing tasks by tailoring deep neural networks but do not explore the intrinsic tension between various deep learning models and distinct WiFi sensing data collected by different devices and CSI tools. It is unclear if the remarkable results of a WiFi sensing research paper come from the deep model design or the WiFi platform. Hence, there still exist some significant gaps between current deep learning and WiFi sensing research: (i) how to customize a deep neural network for a WiFi sensing task by integrating prevailing network modules (\textit{e.g.}, fully-connected layer, convolutional layer, recurrent neural unit, transformer block) into one synergistic framework? (ii) how do the prevailing models perform when they are compared fairly on multiple WiFi sensing platforms and data modalities? (iii) how to achieve a trade-off between recognition accuracy and efficiency?

To answer these questions, we propose SenseFi, a benchmark and model zoo library for WiFi CSI sensing using deep learning. Firstly, we introduce the prevalent deep learning models, including multilayer perceptron (MLP), convolutional neural network (CNN), recurrent neural network (RNN), variants of RNN, CSI transformers, and CNN-RNN, and summarize how they are effective for CSI feature learning and WiFi sensing tasks. Then we investigate and benchmark these models on three WiFi human activity recognition data that consists of both raw CSI data and processed data collected by Intel 5300 CSI tool~\cite{halperin2011tool} and Atheros CSI tool~\cite{xie2015precise,yang2018device}. The accuracy and efficiency of these models are compared and discussed to show their viability for real-world applications. We further investigate how different WiFi sensing tasks can benefit each other by transfer learning, and how unsupervised learning can be used to exploit features without labels, reducing the annotation cost. These features are summarized in Figure~\ref{fig:principle}. All the source codes are written into one library so that the researchers can develop and evaluate their models conveniently.

As such, the contributions are summarized as follows:
\begin{itemize}
    \item We analyze and summarize how the widespread deep learning models in computer vision and natural language processing benefit WiFi sensing in terms of network structure and feature extraction.
    \item We select two public datasets (UT-HAR~\cite{yousefi2017survey} and Widar~\cite{zhang2021widar3}) and collect two new datasets (NTU-Fi HAR and Human-ID) using different CSI platforms, which allows us to benchmark the deep learning methods and evaluate their feasibility for WiFi sensing.
    \item We explore the transfer learning scheme that transfers knowledge across different sensing tasks, and benchmark it across all models.
    \item We investigate the unsupervised learning scheme that contrastively learns the feature extractor without data annotation, and benchmark it across all models.
    \item We develop the \textbf{SenseFi} library and open-source the benchmarking codes. To the best of our knowledge, this is the first work that benchmarks advanced deep models and learning schemes for WiFi sensing, which provides comprehensive and significant evidence and tools for future research.
\end{itemize}


The rest of the paper is organized as follows. Section~\ref{sec:pre} introduces the fundamental knowledge on WiFi sensing and CSI data. Then we introduce the prevalent deep learning models and how they are applied to WiFi sensing in Section~\ref{sec:dl-models}. The empirical study is detailed in Section~\ref{sec:empirical-study}, and then the summaries and discussions are made in Section~\ref{sec:summary}. Finally, the paper is concluded in Section~\ref{sec:conclusion}.

\section{Preliminaries of WiFi Sensing}\label{sec:pre}
\begin{figure*}[t]
	\centering
	\includegraphics[width=1\textwidth]{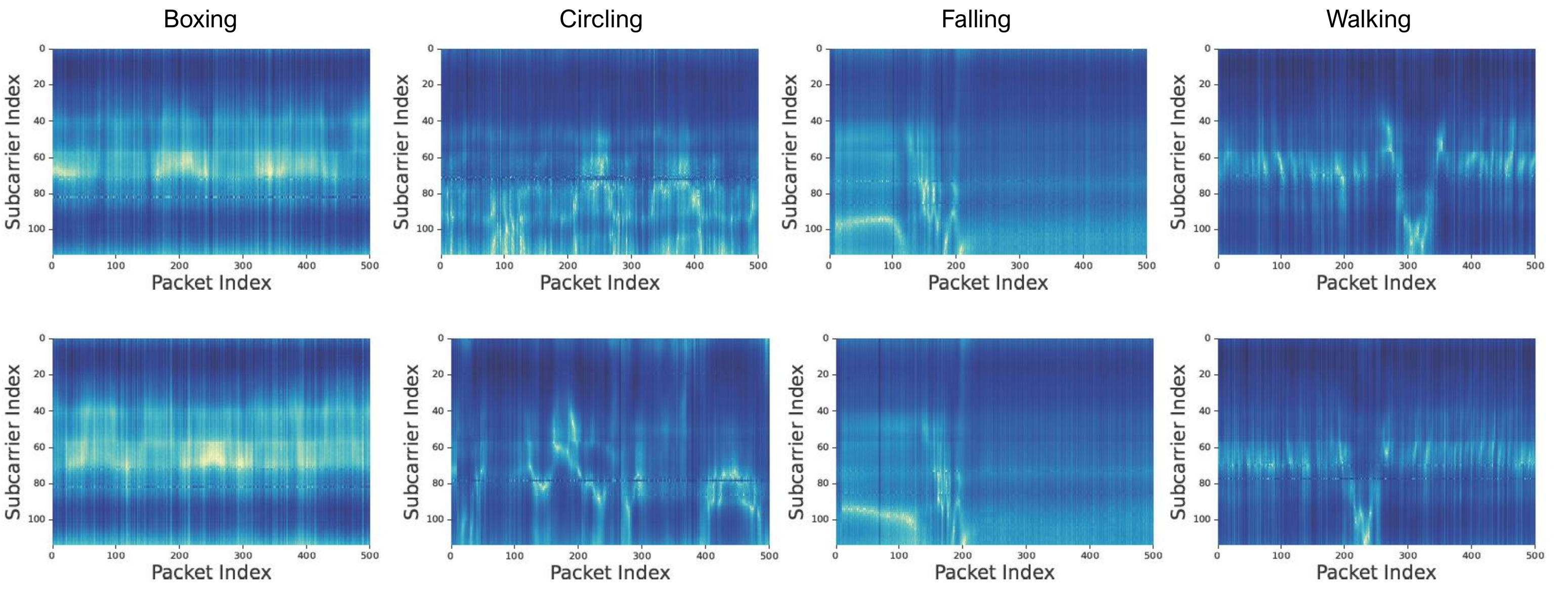}
	\caption{The CSI samples of three human activities in NTU-Fi, collected by Atheros CSI Tool.}\label{fig:csi-samples}
\end{figure*}
\subsection{Channel State Information}
In WiFi communication, channel state information reflects how wireless signals propagate in a physical environment after diffraction, reflections, and scattering, which describes the channel properties of a communication link. For modern wireless communication networks following the IEEE 802.11 standard, Multiple-Input Multiple-Output (MIMO) and Orthogonal Frequency Division Multiplexing (OFDM) at the physical layer contribute to increasing data capacity and better orthogonality in transmission channels affected by multi-path propagation. As a result, current WiFi APs usually have multiple antennas with many subcarriers for OFDM. For a pair of transmitter and receiver antennas, CSI describes the phase shift of multi-path and amplitude attenuation on each subcarrier. Compared to received signal strength, CSI data has better resolutions for sensing and can be regarded as ``WiFi images'' for the environment where WiFi signals propagate. Specifically, the Channel Impulse Response (CIR) $h(\tau)$ of the WiFi signals is defined in the frequency domain:
\begin{equation}
h(\tau)=\sum_{l=1}^{L}\alpha_l e^{j\phi_l} \delta(\tau-\tau_l),
\end{equation}
where $\alpha_l$ and $\phi_l$ denote the amplitude and phase of the $l$-th multi-path component, respectively, $\tau_l$ is the time delay, $L$ denotes the number of multi-paths and $\delta(\tau)$ is the Dirac delta function. To estimate the CIR, the OFDM receiver samples the signal spectrum at subcarrier level in the realistic implementation, which represents amplitude attenuation and phase shift via complex number. In WiFi sensing, the CSI recording functions are realized by specific tools~\cite{halperin2011tool,xie2015precise}. The estimation can be represented by:
\begin{equation}
H_i=||H_i||e^{j \angle H_i}
\end{equation}
where $||H_i||$ and $\angle H_i$ are the amplitude and phase of $i$-th subcarrier, respectively. 

\subsection{CSI Tools and Platforms}
The number of subcarriers is decided by the bandwidth and the tool. The more subcarriers one has, the better resolution the CSI data is. Existing CSI tools include Intel 5300 NIC~\cite{halperin2011tool}, Atheros CSI Tool~\cite{xie2015precise} and Nexmon CSI Tool~\cite{nexmon2019tool}, and many realistic sensing platforms are built on them. The Intel 5300 NIC is the most commonly used tool, which is the first released CSI tool. It can record 30 subcarriers for each pair of antennas running with 20MHz bandwidth. Atheros CSI Tool increases the CSI data resolution by improving the recording CSI to 56 subcarriers for 20MHz and 114 subcarriers for 40MHz, which has been widely used for many applications~\cite{zou2018deepsense,yang2018device,yang2018carefi,yang2019learning,yang2022efficientfi}. The Nexmon CSI Tool firstly enables CSI recording on smartphones and Raspberry Pi, and can capture 256 subcarriers for 80MHz. However, past works~\cite{sharma2021passive,schafer2021human} show that their CSI data is quite noisy, and there do not exist common datasets based on Nexmon. In this paper, we only investigate the effectiveness of the deep learning models trained on representative CSI data from the widely-used Intel 5300 NIC and Atheros CSI Tool. 

\subsection{CSI Data Transformation and Cleansing}
In general, the CSI data consists of a vector of complex number including the amplitude and phase. The question is how we process these data for the deep models of WiFi sensing? We summarize the answers derived from existing works:
\begin{enumerate}
    \item \textbf{Only use the amplitude data as input.} As the raw phases from a single antenna are randomly distributed due to the random phase offsets~\cite{liu2020human}, the amplitude of CSI is more stable and suitable for WiFi sensing. A simple denoising scheme is enough to filter the high-frequency noise of CSI amplitudes, such as the wavelet denoising~\cite{yang2018device}. This is the most common practice for most WiFi sensing applications. 
    \item \textbf{Use the CSI difference between antennas for model-based methods.} Though the raw phases are noisy, the phase difference between two antennas is quite stable~\cite{yang2019learning}, which can better reflect subtle gestures than amplitudes. Then the CSI ratio~\cite{zeng2019farsense} is proposed to mitigate the noise by the division operation and thus increases the sensing range. These techniques are mostly designed for model-based solutions as they require clean data for selecting thresholds.
    \item \textbf{Use the processed doppler representation of CSI.} To eliminate the environmental dependency of CSI data, the body-coordinate velocity profile (BVP) is proposed to simulate the doppler feature~\cite{zhang2021widar3} that only reflects human motions.
\end{enumerate}
In our benchmark, as we focus on the learning-based methods, we choose the most common data modality (\textit{i.e.,} amplitude only) and the novel BVP modality that is domain-invariant.

\subsection{How Human Activities Affect CSI}
As shown in Figure~\ref{fig:csi-samples}, the CSI data for human sensing is composed of two dimensions: the subcarrier and the packet number (\textit{i.e.}, time duration). For each packet or timestamp $t$, we have $X_t=N_T\times N_R \times N_{sub}$ where $N_T$, $N_R$ and $N_{sub}$ denote the number of transmitter antennas, receiver antennas and subcarriers per antenna, respectively. This can be regarded as a ``CSI image'' for the surrounding environment at time $t$. Then along with subsequent timestamps, the CSI images form a ``CSI video'' that can describe human activity patterns. To connect CSI data with deep learning models, we summarize the data properties that serve for a better understanding of deep model design:
\begin{enumerate}
    \item \textbf{Subcarrier dimension $\to$ spatial features.} The values of many subcarriers can represent how the signal propagates after diffraction, reflections, and scattering, and thus describe the spatial environment. These subcarriers are seen as an analogy for image pixels, from which \textit{convolutional layers} can extract spatial features~\cite{lecun2015deep}.
    \item \textbf{Time dimension $\to$ temporal features.} For each subcarrier, its temporal dynamics indicate an environmental change. In deep learning, the temporal dynamics are usually modeled by \textit{recurrent neural networks}~\cite{schuster1997bidirectional}.
    \item \textbf{Antenna dimension $\to$ resolution and channel features.} As each antenna captures a different propagation path of signals, it can be regarded as a channel in deep learning that is similar to RGB channels of an image. If only one pair of antennas exists, then the CSI data is similar to a gray image with only one channel. Hence, the more antennas we have, the higher resolution the CSI has. The antenna features should be processed separately in convolutional layers or recurrent neurons.
\end{enumerate}


\begin{table*}[htp]
\centering
\caption{A Survey of Existing Deep Learning Approaches for WiFi Sensing}\label{tab:survey}
\scalebox{0.94}{
    \begin{tabular}{lccccc} \toprule
    Method         & Year & Task   & Model                 & Platform   &  Strategy    \\ \midrule
    \cite{yousefi2017survey}        & 2017 & Human Activity Recognition    &  RNN, LSTM & Intel 5300 NIC &  Supervised learning  \\
    WiCount~\cite{liu2017wicount}                                                   & 2017 & People Counting     & MLP                                                 & Intel 5300 NIC &  Supervised learning  \\
    EI~\cite{jiang2018towards}                                                      & 2018 & Human Activity Recognition    & CNN                                                 & Intel 5300 NIC &  Transfer learning  \\
    CrossSense~\cite{zhang2018crosssense}                                           & 2018 & Human Identification,Gesture Recognition  & MLP                                                 & Intel 5300 NIC &  Transfer Ensemble learning  \\
    \cite{chen2018wifi}                                                                  & 2018 & Human Activity Recognition    & LSTM                                                & Intel 5300 NIC &  Supervised learning  \\
    DeepSense~\cite{zou2018deepsense}                                               & 2018 & Human Activity Recognition    & CNN-LSTM                                            & Atheros CSI Tool        &  Supervised learning  \\
    WiADG~\cite{zou2018robust}                                                      & 2018 & Gesture Recognition     & CNN                                                 & Atheros CSI Tool        &  Transfer learning  \\
    WiSDAR~\cite{wang2018spatial}                                                   & 2018 & Human Activity Recognition    & CNN-LSTM                                            & Intel 5300 NIC &  Supervised learning  \\
    WiVi~\cite{zou2019wificv}                                                       & 2019 & Human Activity Recognition    & CNN                                                 & Atheros CSI Tool        &  Supervised learning  \\
    SiaNet~\cite{yang2019learning}                                                  & 2019 & Gesture Recognition     & CNN-LSTM                                            & Atheros CSI Tool        &  Few-Shot learning  \\
    CSIGAN~\cite{xiao2019csigan}                                                  & 2019 & Gesture Recognition     & CNN, GAN                                           & Atheros CSI Tool        &  Semi-Supervised learning  \\
    DeepMV~\cite{xue2020deepmv}                                                     & 2020 & Human Activity Recognition    & CNN (Attention)                                     & Intel 5300 NIC &  Supervised learning  \\
    WIHF~\cite{li2020wihf}& 2020 & Gesture Recognition    & CNN-GRU                                     & Intel 5300 NIC &  Supervised learning  \\
    DeepSeg~\cite{xiao2020deepseg}                                                  & 2020 & Human Activity Recognition    & CNN                                                 & Intel 5300 NIC &  Supervised learning  \\
    \cite{sheng2020deep}                                                                 & 2020 & Human Activity Recognition    & CNN-LSTM                                            & Intel 5300 NIC &  Supervised learning  \\
    \cite{schafer2021human} & 2021 & Human Activity Recognition & LSTM                                                 & Nexmon CSI Tool        &  Supervised learning  \\
    \cite{moshiri2021csi} & 2021 & Human Activity Recognition & CNN                                                 & Nexmon CSI Tool        &  Supervised learning  \\
    \cite{ding2021improving}  &  2021 & Human Activity Recognition & CNN & Intel 5300 NIC &  Few-Shot learning  \\
    Widar~\cite{zhang2021widar3}                                                   & 2021 & Human Identification, Gesture Recognition  & CNN-GRU                                             & Intel 5300 NIC &  Supervised learning  \\
    WiONE~\cite{gu2021wione}                                                   & 2021 & Human Identification  & CNN                                             & Intel 5300 NIC &  Few-Shot learning  \\
    \cite{ma2021location}                                                                & 2021 & Human Activity Recognition    & CNN, RNN, LSTM                                        & Intel 5300 NIC &  Supervised learning  \\
    THAT~\cite{li2021two}                                                           & 2021 & Human Activity Recognition    & Transformers                                        & Intel 5300 NIC &  Supervised learning  \\
    WiGr~\cite{zhang2021wifi} & 2021 & Gesture Recognition     & CNN-LSTM                                                 &    Intel 5300 NIC     &  Supervised learning  \\
    MCBAR~\cite{wang2021multimodal} & 2021 & Human Activity Recognition    & CNN, GAN                                                 & Atheros CSI Tool        &  Semi-Supervised learning  \\
    CAUTION~\cite{wang2022caution}                                                  & 2022 & Human Identification     & CNN                                                 & Atheros CSI Tool        &  Few-Shot learning  \\
    CTS-AM~\cite{ding2022wi}                                                        & 2022 & Human Activity Recognition    & CNN (Attention)                                     & Intel 5300 NIC &  Supervised learning  \\
    WiGRUNT~\cite{gu2022wigrunt}                                                        & 2022 & Gesture Recognition    & CNN (Attention)          & Intel 5300 NIC &  Supervised learning  \\
    \cite{zhuravchak2022human} & 2022 & Human Activity Recognition & LSTM                                                 & Nexmon CSI Tool        &  Supervised learning  \\
    EfficientFi~\cite{yang2022efficientfi}                                          & 2022 & Human Activity Recognition, Human Identification & CNN                                                 & Atheros CSI Tool        &  Multi-task Supervised learning  \\
    RobustSense~\cite{yang2022robustsense}                                          & 2022 & Human Activity Recognition, Human Identification & CNN                                                 & Atheros CSI Tool        &  Supervised learning  \\
    AutoFi~\cite{yang2022autofi}                                                    & 2022 & Human Activity Recognition, Human Identification & CNN-MLP                                             & Atheros CSI Tool       &  Unsupervised learning  \\ \bottomrule
    \end{tabular}
    }
\end{table*}

\section{Deep Learning Models for WiFi Sensing}\label{sec:dl-models}
Deep learning enables models composed of many processing layers to learn representations of data, which is a branch of machine learning~\cite{lecun2015deep}. Compared to classic statistical learning that mainly leverages handcrafted features designed by humans with prior knowledge~\cite{jordan2015machine}, deep learning aims to extract features automatically by learning massive labeled data and optimizing the model by back-propagation~\cite{lecun1988theoretical}. The theories of deep learning were developed in the 1980s but they were not attractive due to the need of enormous computational resources. With the development of graphical processing units (GPUs), deep learning techniques have become affordable, and has been widely utilized in computer vision~\cite{voulodimos2018deep}, natural language processing~\cite{otter2020survey}, and interdisciplinary research~\cite{chen2021deep}. 

A standard classification model in deep learning is composed of a feature extractor and a classifier. The classifier normally consists of several fully-connected layers that are able to learn a good decision boundary, while the design of the feature extractor is the key to the success. Extensive works explore a large number of deep architectures for feature extractors~\cite{liu2017survey}, and each of them has specific advantages for some type of data. The deep learning models for WiFi sensing are built on these prevailing architectures to extract patterns of human motions~\cite{zou2018deepsense}. We summarize the latest works on deep models for WiFi sensing in Table~\ref{tab:survey}, and it is observed that the networks of these works are quite similar.

\begin{figure*}[h]
	\centering
	\includegraphics[width=0.8\textwidth]{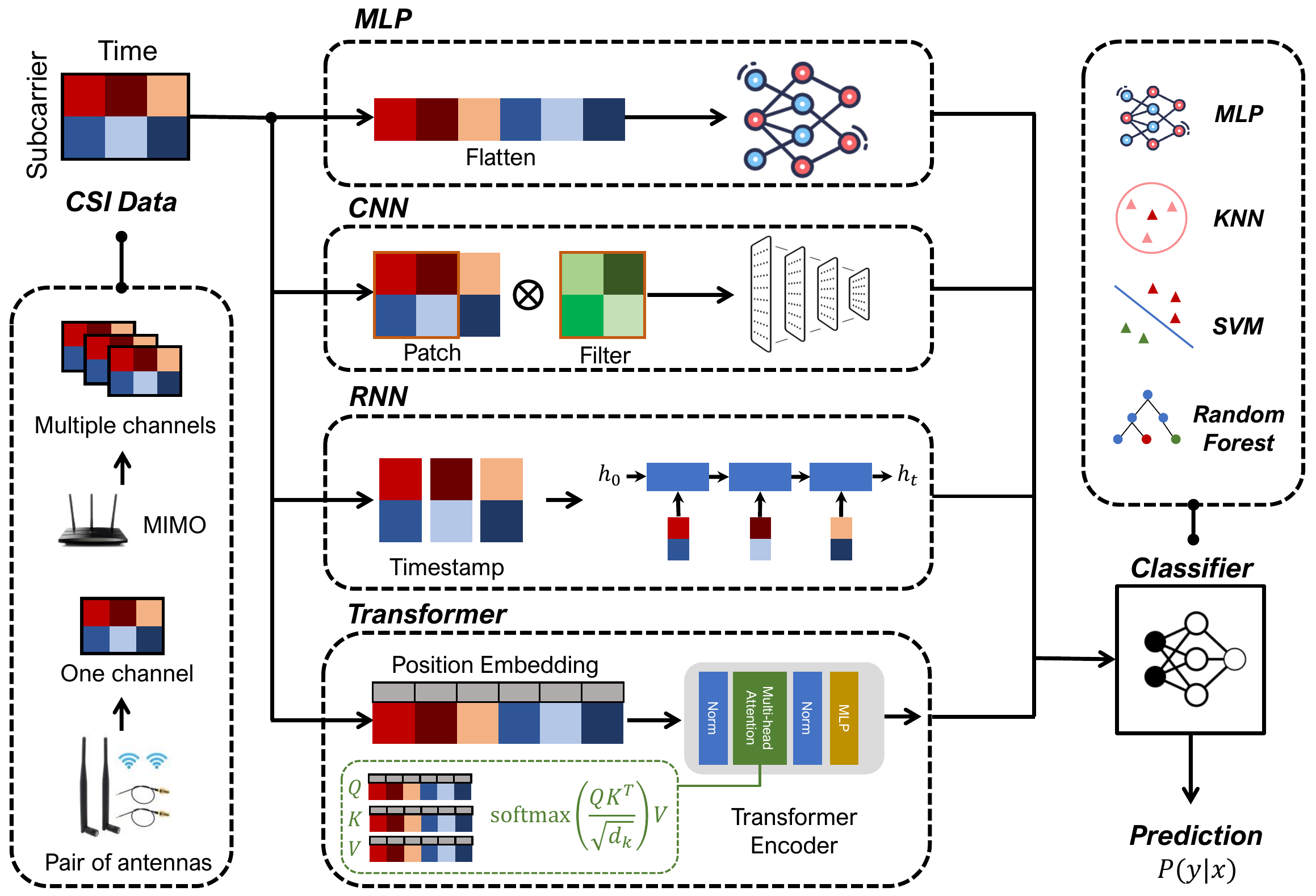}
	\caption{The illustration of how CSI data is processed by MLP, CNN, RNN and Transformer.}
	\label{fig:framework}
\end{figure*}

In the following, we introduce these key architectures and how they are applied to WiFi sensing tasks. To better instantiate these networks, we firstly formulate the normal WiFi CSI sensing task. The CSI data is defined as $x\in \mathbb{R}^{N_a\times N_s\times T}$ where $N_a$ denotes the number of antennas, $N_s$ denotes the number of subcarriers, and $T$ denotes the duration. The CSI data of each pair of antenna is relatively independent, and thus regarded as one channel of CSI, such as the RGB channels of image data. The deep learning model $f(\cdot)$ aims to map the data to the corresponding label: $y=f(x)$, according to different tasks. Denote $\Phi_i(\cdot)$ and $z_i$ as the $i$-th layer of the deep model and the feature of the $i$-th layer. After feature extraction, the classifier is trained to seek a decision boundary in the feature space. In deep learning, the feature extractor is the key module that reduces the feature dimension while preserving the manifold~\cite{lecun2015deep}. With a discriminative feature space, the choices of classifiers are flexible, which can be deep classifier (e.g., Multilayer Perceptron) or traditional classifiers (e.g., K-Nearest Neighbors, Support Vector Machine, and Random Forest). Apart from the illustration, we visualize the intuition of how to feed these CSI data into various networks in Figure~\ref{fig:framework}.

\subsection{Multilayer Perceptron}
Multilayer perceptron (MLP)~\cite{gardner1998artificial} is one of the most classic architectures and has played the classifier role in most deep classification networks. It normally consists of multiple fully-connected layers followed by activation functions. The first layer is termed the input layer that transforms the input data into the hidden latent space, and after several hidden layers, the last layer maps the latent feature into the categorical space. Each layer is calculated as 
\begin{equation}
    \Phi_i(z_{i-1})=\sigma(W_i z_{i-1}),
\end{equation}
where $W_i$ is the parameters of $\Phi_i$, and $\sigma(\cdot)$ is the activation function that aims to increase the non-linearity for MLP. The input CSI has to be flattened to a vector and then fed into the MLP, such that $x\in \mathbb{R}^{N_s T}$. Such a process mixes the spatial and temporal dimensions and damages the intrinsic structure of CSI data. Despite this, the MLP can still work with massive labeled data, because the MLP has a fully-connected structure with a large number of parameters, yet leading to slow convergence and huge computational costs. Therefore, though the MLP shows satisfactory performance, stacking many layers in MLP is not common for feature learning, which makes MLP usually serve as a classifier. In WiFi sensing, MLP is commonly utilized as a classifier~\cite{zou2018deepsense,wang2018spatial,zou2019wificv,zhang2018crosssense,jiang2018towards,zhang2021widar3}.

\subsection{Convolutional Neural Network}
Convolutional neural network (CNN) was firstly proposed for image recognition tasks by LeCun~\cite{lecun1998gradient}. It addresses the drawbacks of MLP by weight sharing and spatial pooling. CNN models have achieved remarkable performances in classification problems of 2D data in computer vision~\cite{khan2020survey,wang2019kervolutional} and sequential data in speech recognition~\cite{abdel2012applying} and natural language processing~\cite{yin2017comparative}. CNN learns features by stacking convolutional kernels and spatial pooling operations. The convolution operation refers to the dot product between a filter $\mathbf{k}\in\mathbb{R}^d$ and an input vector $\mathbf{v}\in\mathbb{R}^d$, defined as follows:
\begin{equation}
    \mathbf{k} \otimes \mathbf{v}=\sigma(\mathbf{k}^T \mathbf{v}).
\end{equation}
The pooling operation is a down-sampling strategy that calculates the maximum (max pooling) or mean (average pooling) inside a kernel. The CNNs normally consist of several convolutional layers, max-pooling layers, and the MLP classifier. Generally speaking, increasing the depth of CNNs can lead to better model capacity. Nevertheless, when the depth of CNN is too large (\textit{e.g.}, greater than 20 layers), the gradient vanishing problem leads to degrading performance. Such degradation is addressed by ResNet~\cite{he2016deep}, which uses the residual connections to reduce the difficulty of optimization.

In WiFi sensing, the convolution kernel can operate on a 2D patch of CSI data (\textit{i.e.}, Conv1D) that includes a spatial-temporal feature, or on a 1D patch of each subcarrier of CSI data (\textit{i.e.}, Conv2D). For Conv2D, a 2D convolution kernel $\mathbf{k}_{2D}\in\mathbb{R}^{h\times w}$ operates on all patches of the CSI data via the sliding window strategy to obtain the output of the feature map, while the Conv1D only extracts the spatial feature along the subcarrier dimension. The Conv2D can be applied independently as it considers both spatial and temporal features, while the Conv1D is usually used with other temporal feature learning methods. To enhance the capacity of CNN, multiple convolution kernels with a random initialization process are used. The advantages of CNNs for WiFi sensing consist of fewer training parameters and the preservation of the subcarrier and time dimension in CSI data. However, the disadvantage is that CNN has an insufficient receptive field due to the limited kernel size and thus fails to capture the dependencies that exceed the kernel size. Another drawback is that CNN stack all the feature maps of kernels equally, which has been revamped by an attention mechanism that assigns different weights in the kernel or spatial level while stacking features. For CSI data, due to the varying locations of human motions, the patterns of different subcarriers should have different importance, which can be depicted by spatial attention in CTS-AM~\cite{ding2022wi}. More attention techniques have been successfully developed to extract temporal-level, antenna-level and subcarrier-level features for WiFi sensing~\cite{xue2020deepmv,9275362,9721516}.

\subsection{Recurrent Neural Network}
Recurrent neural network (RNN) is one of the deepest network architectures that can memorize arbitrary-length sequences of input patterns. The unique advantage of RNN is that it enables multiple inputs and multiple outputs, which makes it very effective for time sequence data, such as video~\cite{yang2018deep} and CSI~\cite{zou2018deepsense,8918311,8761445}. Its principle is to create internal memory to store historical patterns, which are trained via back-propagation through time~\cite{lipton2015rnnsurvey}.

For a CSI sample $x$, we denote a CSI frame at the $t$ as $x_t \in \mathbb{R}^{N_s}$. The vanilla RNN uses two sharing matrices $W_x,W_h$ to generate the hidden state $h_t$:
\begin{equation}
    h_t = \sigma(W_x x_t+W_h h_{t-1}),
\end{equation}
where the activation function $\sigma(\cdot)$ is usually Tanh or Sigmoid functions. RNN is designed to capture temporal dynamics, but it suffers from the vanishing gradient problem during back-propagation and thus cannot capture long-term dependencies of CSI data. 

\subsection{Variants of RNN (LSTM)}
To tackle the problem of long-term dependencies of RNN, Long-short term memory (LSTM)~\cite{hochreiter1997long} is proposed by designing several gates with varying purposes and mitigating the gradient instability during training. The standard LSTM sequentially updates a hidden sequence by a memory cell that contains four states: a memory state $c_t$, an output gate $o_t$ that controls the effect of output, an input gate $i_t$ and a forget gate $f_t$ that decides what to preserve and forget in the memory. The LSTM is parameterized by weight matrices $W_i,W_f,W_c,W_o,U_i,U_f,U_c,U_o$ and biases $b^i,b^f,b^c,b^o$, and the whole update is performed at each $t \in \{ 1,...,T\}$:
\begin{align}
& i_t=\sigma(W_i x_t + U_i h_{t-1}+b^i),   \\
& f_t=\sigma(W_f x_t + U_f h_{t-1}+b^f),  \\
& \tilde{c_t}=tanh(W_c x_t + U_c h_{t-1}+b^c), \\
& c_t=i_t \odot \tilde{c_t} + f_t \odot c_{t-1}, \\
& o_t=\sigma(W_o x_t + U_o h_{t-1} + b^o), \\
& h_t = o_t \odot tanh(c_t),
\end{align}
where $\sigma$ is a Sigmoid function. 

Apart from the LSTM cell~\cite{9722627,9532548,9384510}, the multi-layer and bi-directional structure further boost the model capacity. The bidirectional LSTM (BiLSTM) model processes the sequence in two directions and concatenates the features of the forward input $\grave{x}$ and backward input $\acute{x}$. It has been proven that BiLSTM shows better results than LSTM in~\cite{chen2018wifi,9641123}.

\subsection{Recurrent Convolutional Neural Network}
Though LSTM addresses the long-term dependency, it leads to a large computation overhead. To overcome this issue, Gated Recurrent Unit (GRU) is proposed. GRU combines the forget gate and input gate into one gate, and does not employ the memory state in LSTM, which simplifies the model but can still capture long-term dependency. GRU is regarded as a simple yet effective version of LSTM. Leveraging the simple recurrent network, we can integrate the Conv1D and GRU to extract spatial and temporal features, respectively. \cite{dua2021multi,chen2019semisupervised} show that CNN-GRU is effective for human activity recognition. In WiFi sensing, DeepSense~\cite{zou2018deepsense} firstly proposes Conv2D with LSTM for human activity recognition. CNN-GRU is also revamped for CSI-based human gesture recognition in Widar~\cite{zhang2021widar3}. SiaNet~\cite{yang2019learning} further proposes Conv1D with BiLSTM for few-shot gesture recognition. As they perform quite similarly, we use CNN-GRU with fewer parameters in this paper for the benchmark.

\subsection{Transformer}
Transformer~\cite{vaswani2017attention} was firstly proposed for NLP applications to extract sequence embeddings by exploiting attention of words, and then it was extended to the computer vision field where each patch is regarded as a word and one image consists of many patches~\cite{dosovitskiy2020image}. The vanilla consists of an encoder and a decoder to perform machine translation, and only the encoder is what we need. The transformer block is composed of a multi-head attention layer, a feed-forward neural network (MLP), and layer normalization. Since MLP has been explained in previous section, we mainly introduce the attention mechanism in this section. For a CSI sample $x$, we first divide it into $P$ patches $x_p \in \mathbb{R}^{h\times w}$, of which each patch has contained spatial-temporal features. Then these patches are concatenated and added by positional embeddings that infer the spatial position of patches, which makes the input matrix $v\in \mathbb{R}^{d_k}$ where $d_k=P\times hw$. This matrix is transformed into three different matrices via linear embedding: the query $Q$, the key $K$, and the value $V$. The self-attention process is calculated by
\begin{equation}
    \text{Attention}(Q,K,V)=\text{softmax}(\frac{Q\cdot K^T}{\sqrt{d_k}})\cdot V.
\end{equation}
Intuitively, such a process calculates the attention of any two patches via dot product, \textit{i.e.}, cosine similarity, and then the weighting is performed with normalization to enhance gradient stability for improved training. Multi-head attention just repeats the self-attention several times and enhances the diversity of attentions. The transformer architecture can interconnect with every patch of CSI, which makes it strong if given sufficient training data, such as THAT~\cite{li2021two}. However, transformer has a great number of parameters that makes the training cost expensive, and enormous labeled CSI data is hard to collect, which makes transformers not really attractive for the supervised learning.

\subsection{Generative Models}
Different from the aforementioned discriminative models that mainly conducts classification, generative models aim to capture the data distribution of CSI. Generative Adversarial Network (GAN)~\cite{goodfellow2014generative} is a classic generative model that learns to generate real-like data via an adversarial game between a generative network and a discriminator network. In WiFi sensing, GAN helps deal with the environmental dependency by generating labeled samples in the new environment from the well-trained environment~\cite{xiao2019csigan,wang2021multimodal}. GAN also inspires domain-adversarial training that enables deep models to learn domain-invariant representations for the training and real-world testing environments~\cite{zou2018robust,yang2021robust,yang2020mind,xu2021partial}. Variational network~\cite{kingma2013auto} is another common generative model that maps the input variable to a multivariate latent distribution. Variational autoencoder learns the data distribution by a stochastic variational inference and learning algorithm~\cite{kingma2013auto}, which has been used in CSI-based localization~\cite{kim2021multiview,chen2020fido} and CSI compression~\cite{yang2022efficientfi}. For instance, EfficientFi~\cite{yang2022efficientfi} leverages the quantized variational model to compress the CSI transmission data for large-scale WiFi sensing in the future.

\begin{figure}[t]
	\centering
	\includegraphics[width=0.45\textwidth]{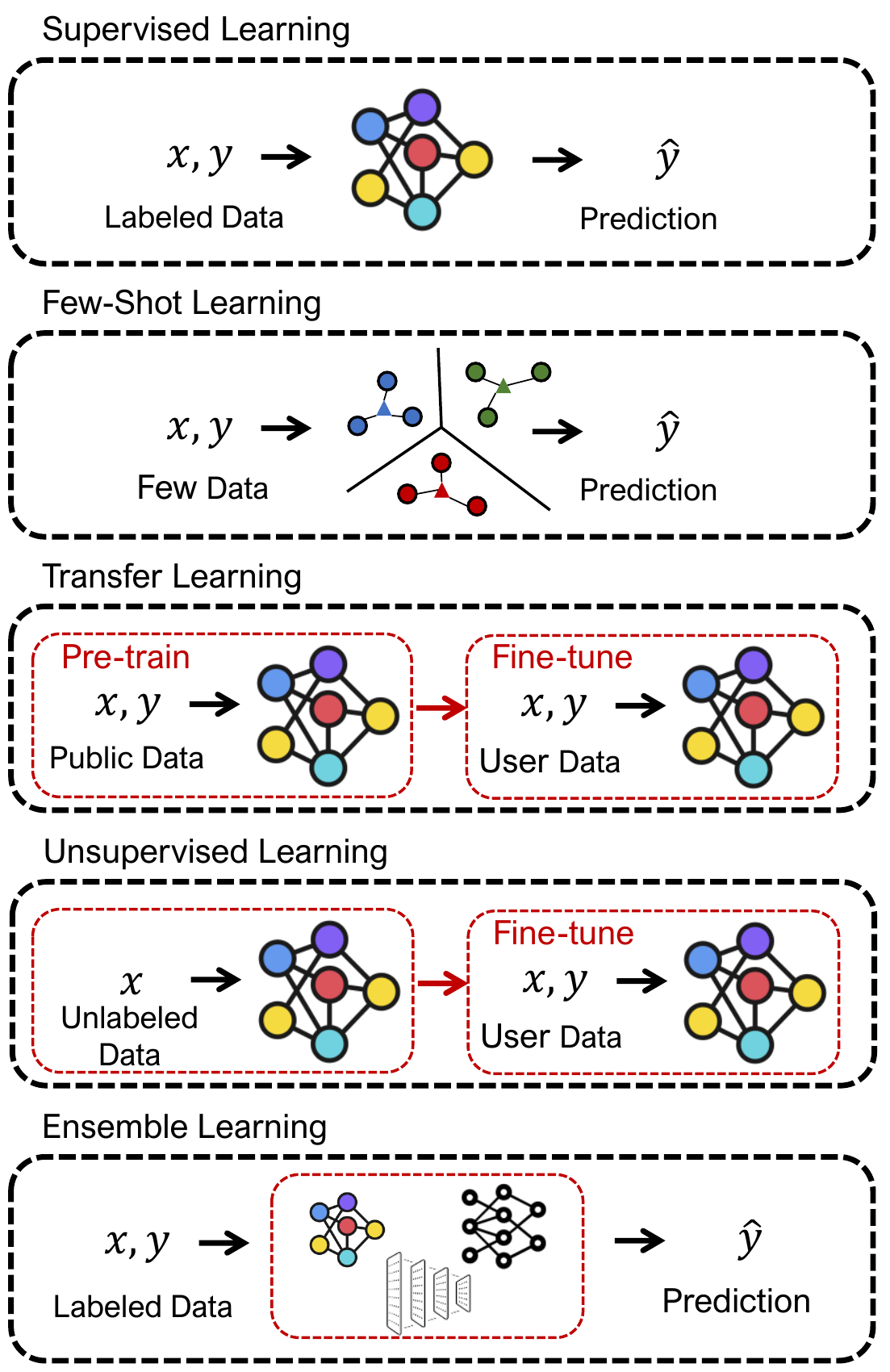}
	\caption{The illustration of the learning strategies.}
	\label{fig:strategies}
\end{figure}

\section{Learning Methods for Deep WiFi Sensing Models}
Traditional training of deep models relies on supervised learning with massive labeled data, but the data collection and annotation is a bottleneck in the realistic WiFi sensing applications. For example, to recognize human gestures, we may need the volunteers to perform gestures for a hundred times, which is not realistic. In this section, as shown in Figure~\ref{fig:strategies}, we illustrate the learning methods and how they contribute to WiFi sensing in the real world.

\textbf{Supervised Learning} is an approach to training deep models using input data that has been labeled for a particular output. It is the most common learning strategy in current WiFi sensing works~\cite{yousefi2017survey,zou2018deepsense,wang2018spatial,zou2019wificv}. They usually adopt cross-entropy loss between the ground truth label and the prediction for model optimization. Though supervised learning is easy to implement and achieves high performance for many tasks, its requirement of tremendous labeled data hinders its pervasive realistic applications.

\textbf{Few-shot Learning} is a data-efficient learning strategy that only utilizes several samples of each category for training. This is normally achieved by contrastive learning or prototypical learning. It is firstly exploited for WiFi sensing in SiaNet~\cite{yang2019learning} that proposes a Siamese network for few-shot learning. Subsequent works~\cite{gu2021wione,wang2022caution} extend prototypical networks from visual recognition to WiFi sensing, also achieving good recognition results. Specially, when only one sample for each class is employed for training, we term it as one-shot learning. As only a few samples are required, few-shot learning contributes to WiFi-based gesture recognition and human identification in practice.

\textbf{Transfer Learning} aims to transfer knowledge from one domain to another domain~\cite{tlsurvey}. When the two domains are similar, we pretrain the model on one domain and fine-tune the model in a new environment, which can lead to significant performance. When the two domains are distinct, such as the different environments of CSI data, the distribution shift hinders the performance so domain adaptation should be adopted. Domain adaptation is a category of semi-supervised learning that mitigates the domain shift for transfer learning. Cross-domain scenarios are quite common in WiFi sensing scenarios since the CSI data is highly dependent on the training environment. Many works have been developed to deal with this problem~\cite{zou2018robust,jiang2018towards,wang2021multimodal,8793019,9763693}.

\textbf{Unsupervised Learning} aims to learn data representations without any labels. Then the feature extractor can facilitate down-streaming tasks by training a specific classifier. From the experience of visual recognition tasks~\cite{grill2020bootstrap}, unsupervised learning can even enforce the model to gain better generalization ability since the model is not dependent on any specific tasks. Current unsupervised learning models are based on self-supervised learning~\cite{wang2021self}. Despite its effectiveness, the unsupervised learning has not been well exploited in WiFi sensing, and only AutoFi is developed to enable model initialization for automatic user setup in WiFi sensing applications ~\cite{yang2022autofi}.

\textbf{Ensemble Learning} uses multiple models to obtain better predictive performance~\cite{sagi2018ensemble}. The ensemble process can operate on feature level or prediction level. Feature-level ensemble concatenates the features from multiple models and one final classifier is trained. Prediction-level ensemble is more common, usually referring to voting or probability addition. Ensemble learning can increase the performance but the computation overhead also explodes by multiple times. CrossSense~\cite{jiang2018towards} develops a mixture-of-experts approach and only chooses the appropriate expert for a specific input, addressing the computation cost. 

In this paper, we empirically explore the effectiveness of supervised learning, transfer learning and unsupervised learning for WiFi CSI data, as they are the most commonly used learning strategies in WiFi sensing applications.

\begin{table*}[t]
\centering
\caption{Statistics of four CSI datasets for our SenseFi benchmarks.}\label{tab:datasets}
\begin{tabular}{l|cccc}
\toprule
Datasets                & UT-HAR~\cite{yousefi2017survey}                                                                                 & Widar~\cite{zhang2021widar3}                                                                                       & NTU-Fi HAR~\cite{yang2022efficientfi}                                                                         & NTU-Fi Human-ID~\cite{wang2021multimodal}                                                                             \\ \midrule
Platform         & Intel 5300 NIC                                                                                        & Intel 5300 NIC                                                                                           & Atheros CSI Tool                                                                                   & Atheros CSI Tool                                                                                  \\ \midrule
Category Number         & 7                                                                                        & 22                                                                                           & 6                                                                                   & 14                                                                                  \\ \midrule
Category Names          & \begin{tabular}[c]{@{}c@{}}Lie down, Fall, Walk, Pick up \\ Run, Sit down, Stand up\end{tabular} & \begin{tabular}[c]{@{}c@{}}Push\&Pull, Sweep, Clap, Slide, \\ 18 types of Draws\end{tabular} & \begin{tabular}[c]{@{}c@{}}Box, Circle, Clean, \\ Fall, Run, Walk\end{tabular}      & Gaits of 14 Subjects                                                                \\ \midrule
Data Size               & \begin{tabular}[c]{@{}c@{}}(3,30,250)\\ (antenna, subcarrier, packet)\end{tabular}       & \begin{tabular}[c]{@{}c@{}}(22,20,20)\\ (time, x\_velocity, y\_velocity)\end{tabular}        & \begin{tabular}[c]{@{}c@{}}(3,114,500)\\ (antenna, subcarrier, packet)\end{tabular} & \begin{tabular}[c]{@{}c@{}}(3,114,500)\\ (antenna, subcarrier, packet)\end{tabular} \\ \midrule
Training Samples & 3977                                                                                     & 34926                                                                                        & 936                                                                                 & 546                                                                                 \\ \midrule
Testing Samples  & 996                                                                                      & 8726                                                                                         & 264                                                                                 & 294              \\ \midrule
Training Epochs  & 200                                                                                      & 100                                                                                         & 30                                                                                 & 30              \\
\bottomrule
\end{tabular}
\end{table*}

\begin{table*}[]
\centering
\caption{Evaluation of deep neural networks (using supervised learning) on four datasets. (\textbf{Bold}: best; \underline{Underline}: 2nd best)}\label{tab:overall}
\scalebox{0.84}{
    \begin{tabular}{l|ccc|ccc|ccc|ccc}
    \toprule
    Dataset     & \multicolumn{3}{c|}{\textbf{UT-HAR}}             & \multicolumn{3}{c|}{\textbf{Widar}}              & \multicolumn{3}{c|}{\textbf{NTU-Fi HAR}}            & \multicolumn{3}{c}{\textbf{NTU-Fi Human-ID}}       \\
    Method      & Acc (\%)  & Flops (M)     & Params (M)     & Acc (\%)  & Flops (M)     & Params (M)     & Acc (\%)  & Flops (M)      & Params (M)     & Acc (\%)  & Flops (M)      & Params (M)     \\ \midrule
    MLP         & 92.00          & 23.17         & 23.170         & 67.24    & 9.15          & 9.150          &  \textbf{99.69}    & 175.24         & 175.240        & 93.91          & 175.24         & 175.240        \\
    CNN-5      & \underline{97.61}    & 31.68         & \underline{0.296}    & \underline{70.19} & 3.38          & 0.299          & 98.70          & \underline{28.24}    & 0.477          & 97.14          & \underline{28.24}    & 0.478          \\
    ResNet18    & \textbf{98.11} & 49.93         & 11.180         & \textbf{71.70}          & 38.39         & 11.250         & 95.31          & 54.19          & 11.180         & 96.42          & 54.19          & 11.190         \\
    ResNet50    & 97.21          & 86.40         & 23.550         & 68.56          & 69.70         & 23.640         & \underline{99.38}          & 90.66          & 23.550         & 92.91          & 90.67          & 23.570         \\
    ResNet101   & 94.99          & 162.58        & 42.570         & 68.71          & 145.87        & 42.660         & 95.31          & 166.83         & 42.570         & 88.40          & 166.85         & 42.590         \\
    RNN         & 83.53          & \textbf{2.51} & \textbf{0.010} & 47.05          & \textbf{0.66} & \textbf{0.031} & 84.64          & \textbf{13.09} & \textbf{0.027} & 89.30          & \textbf{13.09} & \textbf{0.027} \\
    GRU         & 94.18          & \underline{7.60}    & 0.030          & 62.50          & \underline{1.98}    & \underline{0.091}    & 97.66          & 39.39          & 0.079          & \underline{98.96}    & 39.39          & 0.079          \\
    LSTM        & 87.18          & 10.14         & 0.040          & 63.35          & 2.64          & 0.121          & 97.14          & 52.54          & 0.105          & 97.19          & 52.54          & 0.105          \\
    BiLSTM      & 90.19          & 20.29         & 0.080          & 63.43          & 5.28          & 0.240          & \textbf{99.69} & 105.09         & 0.209          & \textbf{99.38} & 105.09         & 0.210          \\
    CNN + GRU   & 96.72          & 39.99         & 1.430          & 63.19          & 3.34          & 0.092          & 93.75          & 48.38          & \underline{0.058}    & 87.48          & 48.39          & \underline{0.058}    \\
    ViT & 96.53          & 273.10        & 10.580         & 67.72          & 9.28          & 0.106          & 93.75          & 501.64         & 1.052          & 76.84          & 501.64         & 1.054          \\ \bottomrule
    \end{tabular}
}
\end{table*}


\section{Empirical Studies of Deep Learning in WiFi Sensing: A Benchmark}\label{sec:empirical-study}
In this section, we conduct an empirical study of the aforementioned deep learning models on WiFi sensing data and firstly provide the benchmarks with open-source codes in \url{http://www.github.com/}. The four datasets are illustrated first, and then we evaluate the deep models on these datasets in terms of three learning strategies. Eventually, some detailed analytics are conducted on the convergence of optimization, network depth, and network selection.

\subsection{Datasets}
We choose two public CSI datasets (UT-HAR~\cite{yousefi2017survey} and Widar~\cite{zhang2021widar3}) collected using Intel 5300 NIC. To validate the effectiveness of deep learning models on CSI data of different platforms, we collect two new datasets using Atheros CSI Tool~\cite{xie2015precise} and our embedded IoT system~\cite{yang2018device}, namely NTU-Fi HAR and NTU-Fi Human-ID. The details and statistics of these datasets are summarized in Table~\ref{tab:datasets}. 

\textbf{UT-HAR}~\cite{yousefi2017survey} is the first public CSI dataset for human activity recognition. It consists of seven categories and is collected via Intel 5300 NIC with 3 pairs of antennas that record 30 subcarriers per pair. All the data is collected in the same environment. However, its data is collected continuously and has no golden labels for activity segmentation. Following existing works~\cite{li2021two}, the data is segmented using a sliding window, inevitably causing many repeated data among samples. Hence, though the total number of samples reaches around 5000, it is a small dataset with intrinsic drawbacks.

\textbf{Widar}~\cite{zhang2021widar3} is the largest WiFi sensing dataset for gesture recognition, which is composed of 22 categories and 43K samples. It is collected via Intel 5300 NIC with $3\times3$ pairs of antennas in many distinct environments. To eliminate the environmental dependencies, the data is processed to the body-coordinate velocity profile (BVP). 

\textbf{NTU-Fi} is our proposed dataset for this benchmark that includes both human activity recognition (\textbf{HAR}) and human identification (\textbf{Human ID}) tasks. Different from UT-HAR and Widar, our dataset is collected using Atheros CSI Tool and has a higher resolution of subcarriers (114 per pair of antennas). Each CSI sample is perfectly segmented. For the HAR dataset, we collect the data in three different layouts. For the Human ID dataset, we collect the human walking gaits in three situations: wearing a T-shirt, a coat, or a backpack, which brings many difficulties. The NTU-Fi data is simultaneously collected in these works~\cite{yang2022efficientfi,wang2022caution} that describe the detailed layouts for data collection.

\subsection{Implementation Details}
We normalize the data for each dataset and implement all the aforementioned methods using the PyTorch framework~\cite{paszke2019pytorch}. To ensure the convergence, we train the UT-HAR, Widar, and NTU-Fi for 200, 100, and 30 epochs, respectively, for all the models except RNN. As the vanilla RNN is hard to converge due to the gradient vanishing, we train them for two times of the specified epochs. We use the Adam optimizer with a learning rate of 0.001, and the beta of 0.9 and 0.999. We follow the original Adam paper~\cite{kingma2014adam} to set these hyper-parameters. The ratio of training and testing splits is 8:2 for all datasets using stratified sampling.

\subsection{Baselines and Criterion}
We design the baseline networks of MLP, CNN, RNN, GRU, LSTM, BiLSTM, CNN+GRU, and Transformer following the experiences learned from existing works in Table~\ref{tab:survey}. The CNN-5 is modified from LeNet-5~\cite{lecun1998gradient}. We further introduce the series of ResNet~\cite{he2016deep} that have deeper layers. The transformer network is based on the vision transformer (ViT)~\cite{dosovitskiy2020image} so that each patch can contain spatial and temporal dimensions. It is found that given sufficient parameters and reasonable depth of layers, they can converge to more than 98\% in the training split. Since the data sizes of UT-HAR, Widar and NTU-Fi are different, we use a convolutional layer to map them into a unified size, which enables us to use the same network architecture. The specific network architectures for all models are illustrated in the \textit{Appendix}. The hyper-parameters of the networks have been tuned to ensure the satisfactory convergence. To compare the baseline models, we select three classic criteria: accuracy (Acc) that evaluates the prediction ability, floating-point operations (Flops) that evaluates the computational complexity, and the number of parameters (Params) that measures the requirement of GPU memory. As WiFi sensing is usually performed on the edge, the Flops and Params also matter with limited resources. 

\begin{figure}[t]
	\centering
	\includegraphics[width=0.48\textwidth]{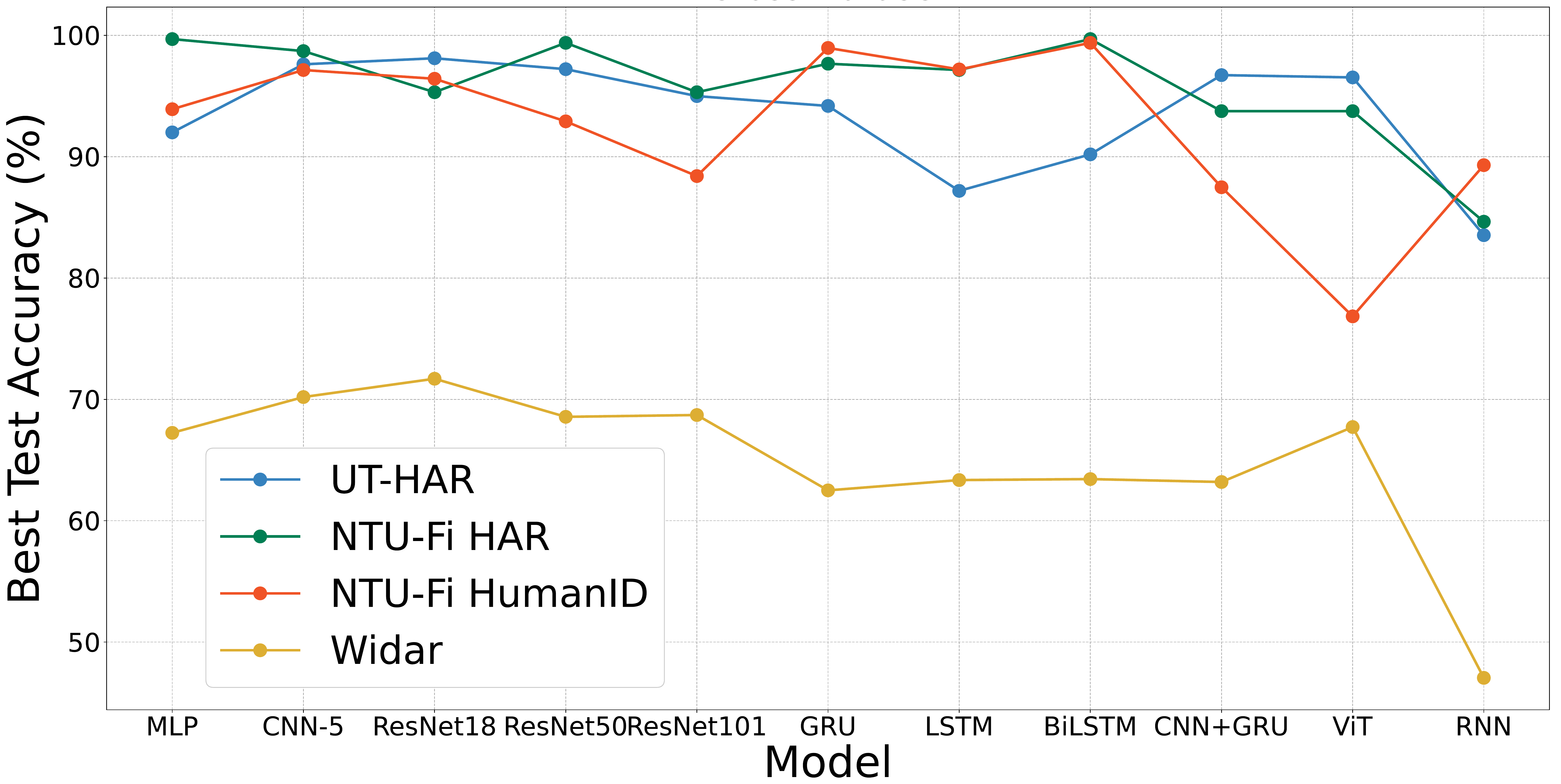}
	\caption{The performance comparison across four datasets.}\label{fig:cross-dataset}
\end{figure}

\subsection{Evaluations of Different Deep Architectures}
\textbf{Overall Comparison.} We summarize the performance of all baseline models in Table~\ref{tab:overall}. On UT-HAR, the ResNet-18 achieves the best accuracy of 98.11\% and the CNN-5 achieves the second best. The shallow CNN-5 can attain good results on all datasets but the deep ResNet-18 fails to generalize on Widar, which will be explained in Section~\ref{sec:exp-analytics}. The BiLSTM yields the best performance on two NTU-Fi benchmarks. To compare these results, we visualize them in Figure~\ref{fig:cross-dataset}, from which we can conclude the observations:
\begin{itemize}
    \item The MLP, CNN, GRU, LSTM, and Transformer can achieve satisfactory results on all benchmarks. 
    \item The MLP, GRU, and CNN show stable and superior performances when they are compared to others.
    \item The very deep networks (\textit{i.e.}, the series of ResNet) perform well on UT-HAR and Widar, but do not perform better than simple CNN on NTU-Fi. The performance does not increase as the number of network layers increases, which is different from visual recognition results~\cite{he2016deep}. Compared to simple CNN-5, the improvement margin is quite limited.
    \item The RNN is worse than LSTM and GRU.
    \item The transformer cannot work well when only limited training data is available in NTU-Fi Human-ID.
    \item The models show inconsistent performances on different datasets, as the Widar dataset is much more difficult.
\end{itemize}

\textbf{Computational Complexity.} The Flops value shows the computational complexity of models in Table~\ref{tab:overall}. The vanilla RNN has low complexity but cannot perform well. The GRU and CNN-5 are the second-best models and simultaneously generate  good results. It is also noteworthy that the ViT (transformer) has a very large computational complexity as it is composed of many MLPs for feature embedding. Since its performance is similar to that of CNN, MLP, and GRU, the transformer is not suitable for supervised learning tasks in WiFi sensing.

\textbf{Model Parameters.} The number of model parameters determines how many GPU memories are occupied during inference. As shown in Table~\ref{tab:overall}, the vanilla RNN has the smallest parameter size and then is followed by the CNN-5 and CNN-GRU. The parameter sizes of CNN-5, RNN, GRU, LSTM, BiLSTM, and CNN-GRU are all small and acceptable for model inference in the edge. Considering both the Params and Acc, CNN-5, GRU, BiLSTM, and CNN-GRU are good choices for WiFi sensing. Though the model parameters can be reduced by model pruning~\cite{chen2019metaquant}, quantization~\cite{chen2019cooperative} or fine-tuning the hyper-parameters, here we only evaluate the pure models that have the minimum parameter sizes to converge in the training split.

\subsection{Evaluations of Learning Schemes}
Apart from supervised learning, other learning schemes are also useful for realistic applications of WiFi sensing. Here we evaluate two prevailing learning strategies on these models.

\begin{table}[htp]
\centering
\caption{Evaluations on Transfer Learning}\label{tab:transfer_learning}
\resizebox{0.9\linewidth}{!}{
    \begin{tabular}{l|ccc}
    \toprule
    Method    & Accuracy (\%)    & Flops (M)      & Params (M)     \\ \midrule
    MLP       & 84.46    & 175.24         & 175.240        \\
    CNN-5     & \textbf{96.35} & \underline{28.24}    & 0.478          \\
    ResNet18  & \underline{85.94}          & 54.19          & 11.190         \\
    ResNet50  & 79.21          & 90.67          & 23.570         \\
    ResNet101 & 68.88          & 166.85         & 42.590         \\
    RNN       & 57.84          & \textbf{13.09} & \textbf{0.027} \\
    GRU       & 75.89          & 39.39          & 0.079          \\
    LSTM      & 71.98          & 52.54          & 0.105          \\
    BiLSTM      & 80.20          & 105.09          & 0.210          \\
    CNN + GRU & 51.73          & 48.39          & \underline{0.059}    \\
    ViT       & 66.20           & 501.64         & 1.054         \\ \bottomrule
    \end{tabular}
}
\end{table}
\begin{figure}[t]
	\centering
    	\includegraphics[width=0.235\textwidth, angle=0]{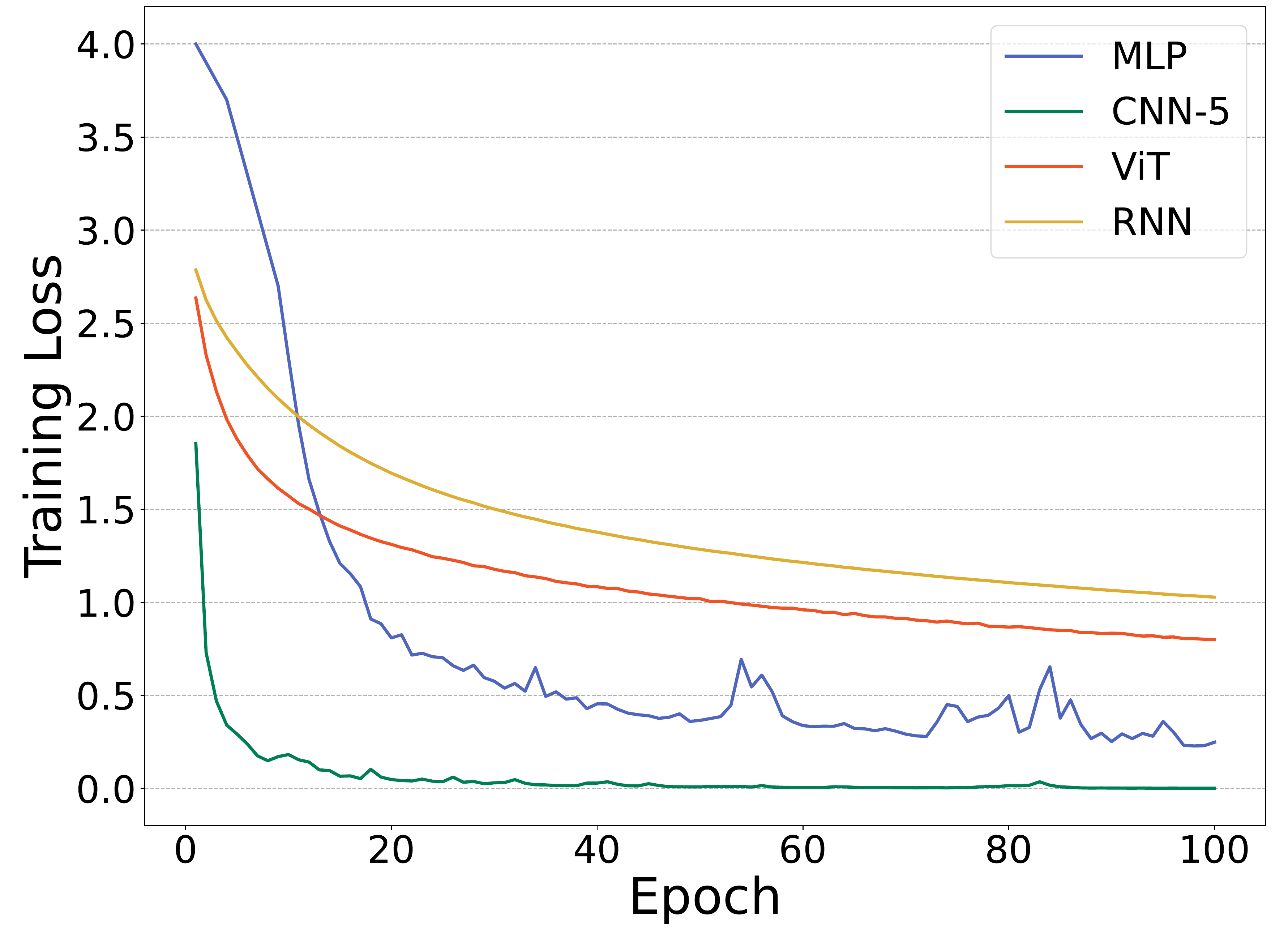}
    	\includegraphics[width=0.235\textwidth, angle=0]{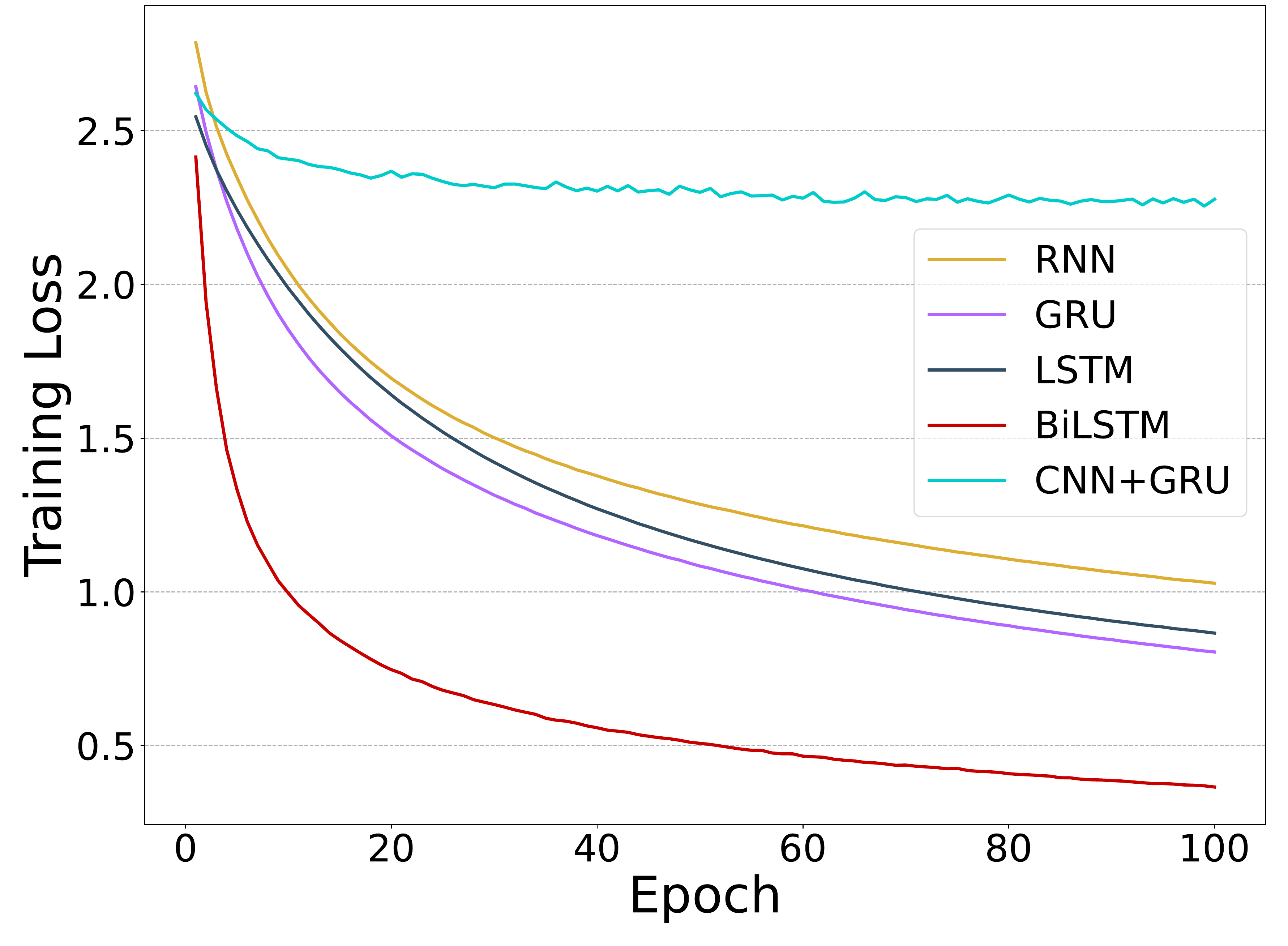}
	\caption{\label{fig:transfer-loss}The training losses of all baseline models on NTU-Fi Human-ID with pre-trained parameters of NTU-Fi HAR.}
\end{figure}

\textbf{Evaluations on Transfer Learning}
The transfer learning experiments are conducted on NTU-Fi. We transfer the model from the HAR to Human-ID by pre-training the model in HAR (whole dataset) and then fine-tuning a new classifier in Human-ID (training split). This simulates the situation when we train the model using massive labeled data collected in the lab, and then use a few data to realize customized tasks for users. The human activities in HAR and human gaits in Human-ID are composed of human motions, and thus the feature extractor should learn to generalize across these two tasks. We evaluate this setting for all baseline models and the results are shown in Table~\ref{tab:transfer_learning}. It is observed that the CNN feature extractor has the best transferability, achieving 96.35\% on the Human-ID task. Similar to CNN, the MLP and BiLSTM also have such capacity. However, the RNN, CNN+GRU, and ViT only achieve 57.84\%, 51.73\%, and 66.20\%, which demonstrates their weaker capacity for transfer learning. This can be caused by the overfitting phenomenon, such as the simple RNN that only memorizes the specific patterns for HAR but cannot recognize the new patterns. This can also be caused by the mechanism of feature learning. For example, the transformer (ViT) learns the connections of local patches by self-attention, but such connections are different between HAR and Human-ID. Recognizing different activities relies on the difference of a series of motions, but most human gaits are so similar that only subtle patterns can be an indicator for gait identification.

\textbf{Evaluations on Unsupervised Learning}
We further exploit the effectiveness of unsupervised learning for CSI feature learning. We follow the AutoFi~\cite{yang2022autofi} to construct two parallel networks and adopt the KL-divergence, mutual information, and kernel density estimation loss to train the two networks only using the CSI data. After unsupervised learning, we train the independent classifier based on the fixed parameters of the two networks. All the backbone networks are tested using the same strategy: unsupervised training on NTU-Fi HAR and supervised learning on NTU-Fi Human-ID. The evaluation is conducted on Human-ID, and the results are shown in Table~\ref{tab:unsupervised_learning}. It is shown that CNN achieves the best accuracy of 97.62\% that is followed by MLP and ViT. The results demonstrate that unsupervised learning is effective for CSI data. It yields better cross-task evaluation results than those of transfer learning, which demonstrates that unsupervised learning helps learn features with better generalization ability. CNN and MLP-based networks are more friendly for unsupervised learning of WiFi CSI data. 

\begin{table}[htp]
\centering
\caption{Evaluations on Unsupervised Learning}\label{tab:unsupervised_learning}
\resizebox{0.9\linewidth}{!}{
    \begin{tabular}{lcccc}
    \toprule
    {\multirow{2}{*}{Method}} & \multicolumn{2}{c}{Accuracy (\%)}                 & \multirow{2}{*}{Flops (M)} & \multirow{2}{*}{Params (M)} \\
    \multicolumn{1}{c}{}                        & \multicolumn{1}{l}{classifier1} & classifier2    &                            &                             \\ \midrule
    MLP                                         & \underline{90.48}                     & \underline{89.12}    & 175.24                     & 175.240                     \\
    CNN-5                                       & \textbf{96.26}                  & \textbf{97.62} & \underline{28.24}                & 0.478                       \\
    ResNet18                                    & 85.03                           & 82.99          & 54.19                      & 11.190                      \\
    ResNet50                                    & 47.28                           & 45.58          & 90.67                      & 23.570                      \\
    ResNet101                                   & 36.05                           & 35.37          & 166.85                     & 42.590                      \\
    RNN                                         & 53.74                           & 51.36          & \textbf{13.09}             & \textbf{0.027}              \\
    GRU                                         & 65.99                           & 64.63          & 39.39                      & 0.079                       \\
    LSTM                                        & 53.06                           & 55.10          & 52.54                      & 0.105                       \\
    BiLSTM                                      & 51.36                           & 55.78          & 105.09                     & 0.210                       \\
    CNN + GRU                                   & 50.34                           & 53.40          & 48.39                      & \underline{0.059}                 \\
    ViT                                         & 78.91                           & 84.35          & 501.64                     & 1.054                      \\ \bottomrule
    \end{tabular}
}
\end{table}

\begin{figure*}[t]
	\centering
	\subfigure[UT-HAR]{
    	\includegraphics[width=0.235\textwidth, angle=0]{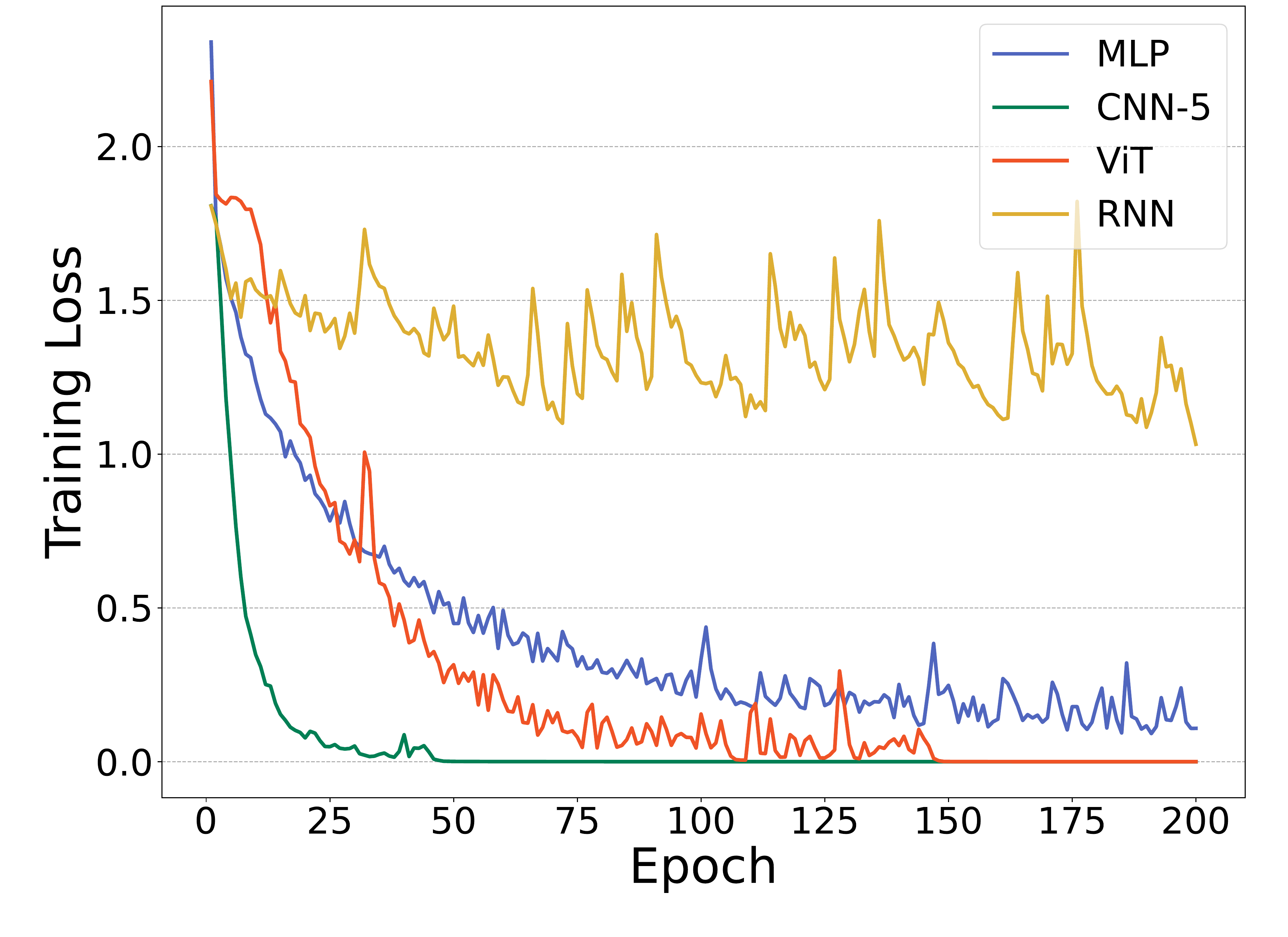}}
	\subfigure[Widar]{
    	\includegraphics[width=0.235\textwidth, angle=0]{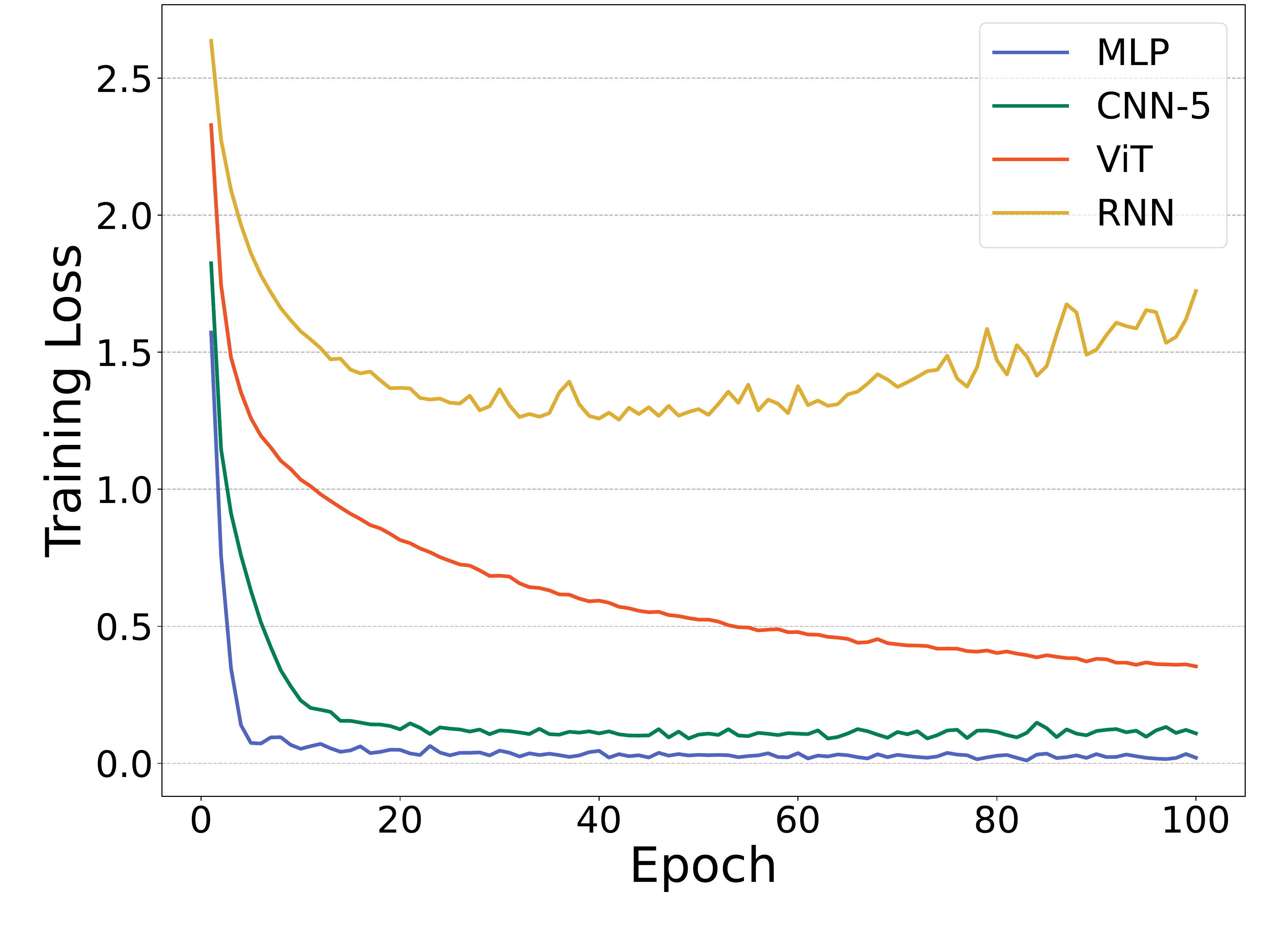}}
	\subfigure[NTU-Fi HAR]{
    	\includegraphics[width=0.235\textwidth, angle=0]{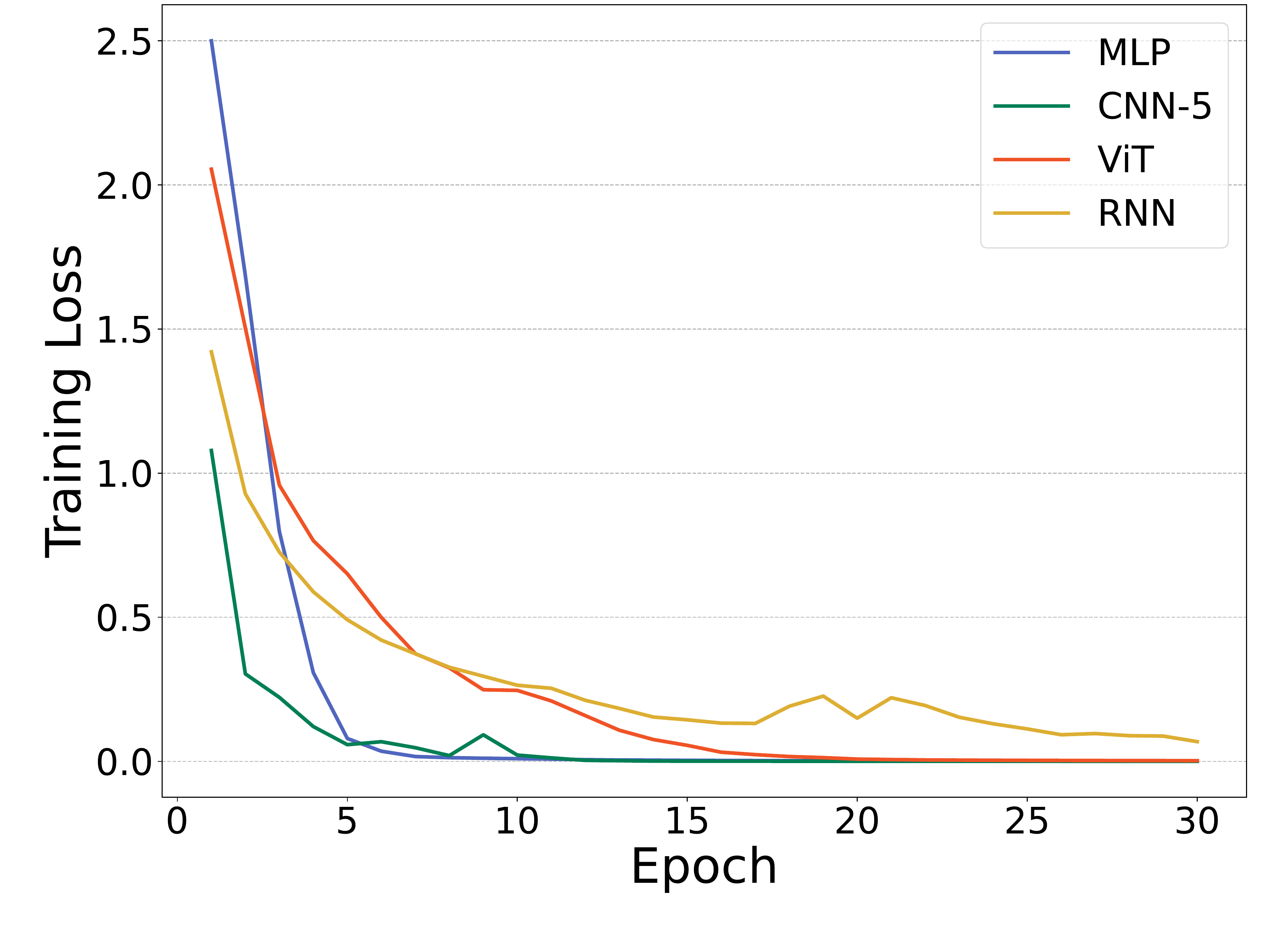}\label{subfig:NTU-HAR-GP1}}
	\subfigure[NTU-Fi Human-ID]{
    	\includegraphics[width=0.235\textwidth, angle=0]{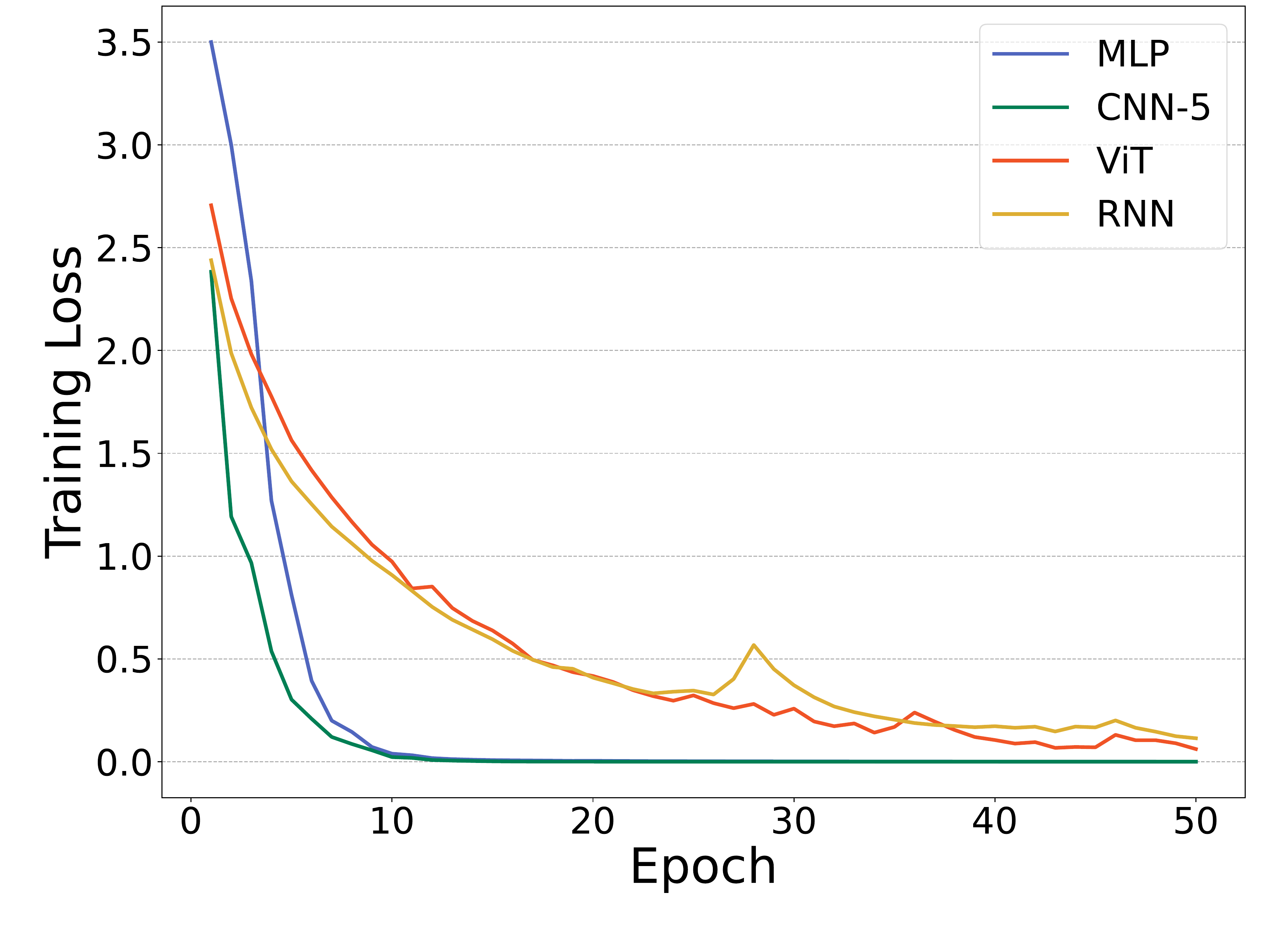}}
	\caption{\label{fig:training-loss}The training losses of MLP, CNN, Transformer, RNN for the four datasets.}
\end{figure*}

\begin{figure*}[t]
	\centering
	\subfigure[UT-HAR]{
    	\includegraphics[width=0.235\textwidth, angle=0]{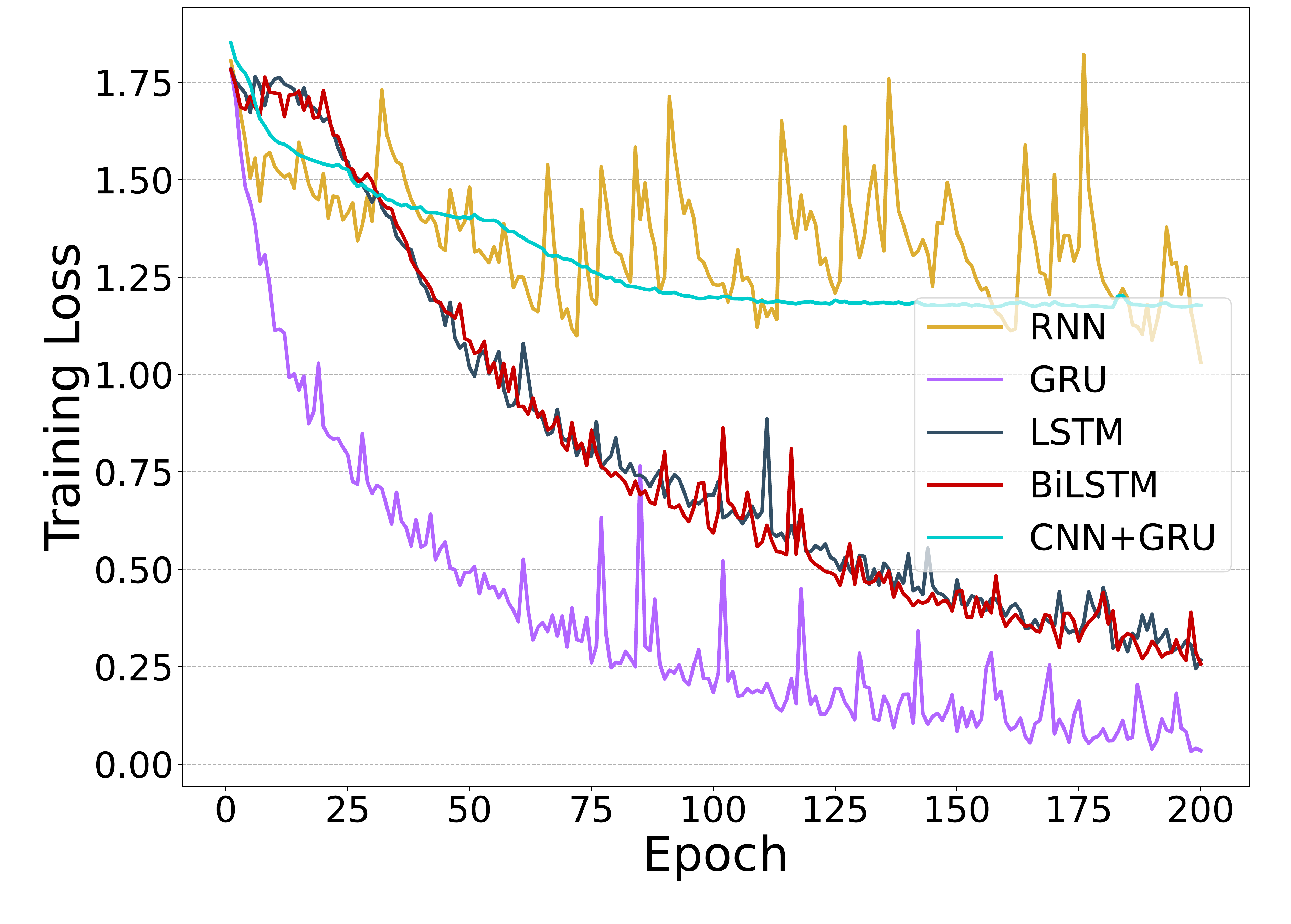}}
	\subfigure[Widar]{
    	\includegraphics[width=0.235\textwidth, angle=0]{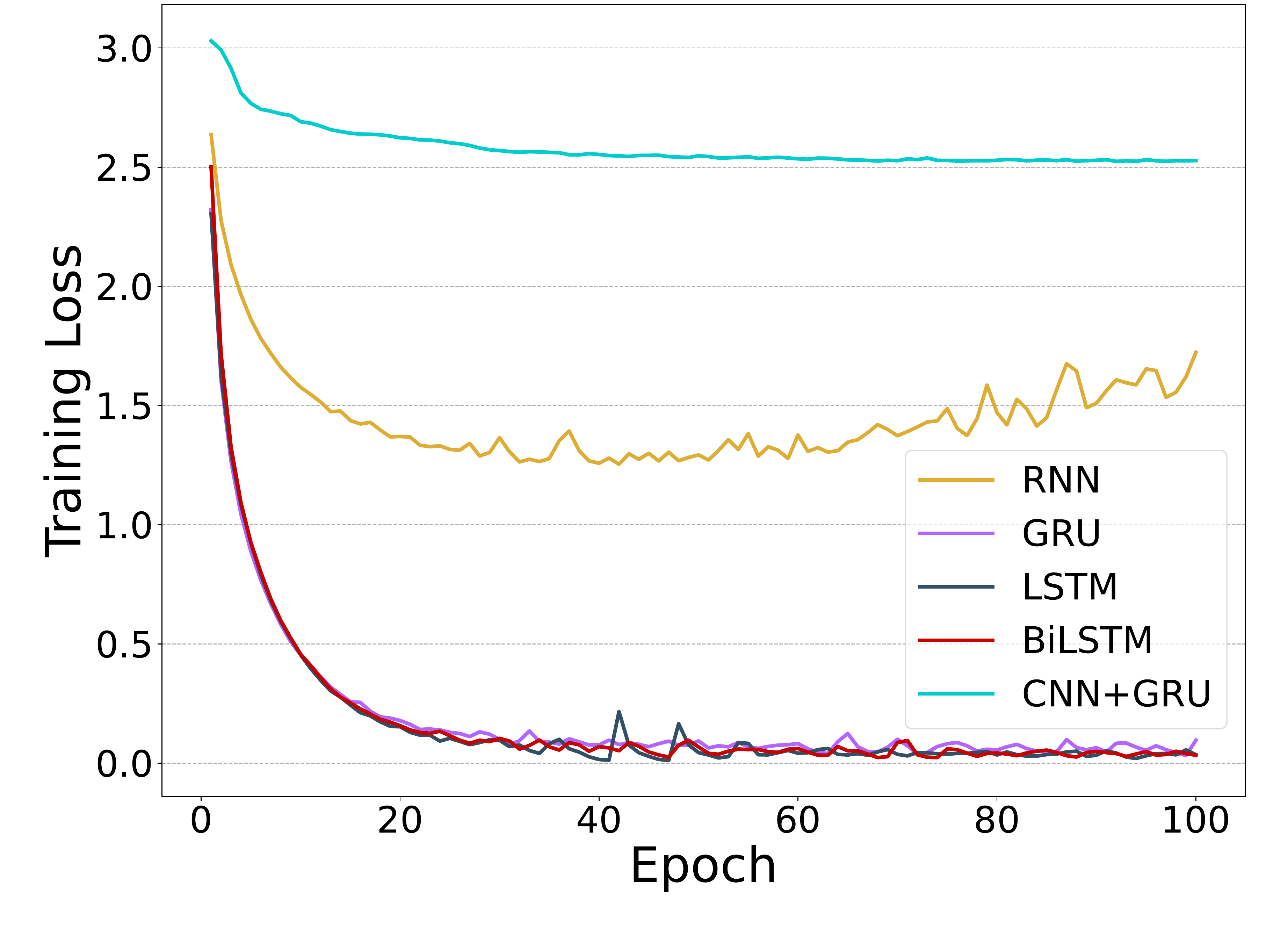}}
	\subfigure[NTU-Fi HAR]{
    	\includegraphics[width=0.235\textwidth, angle=0]{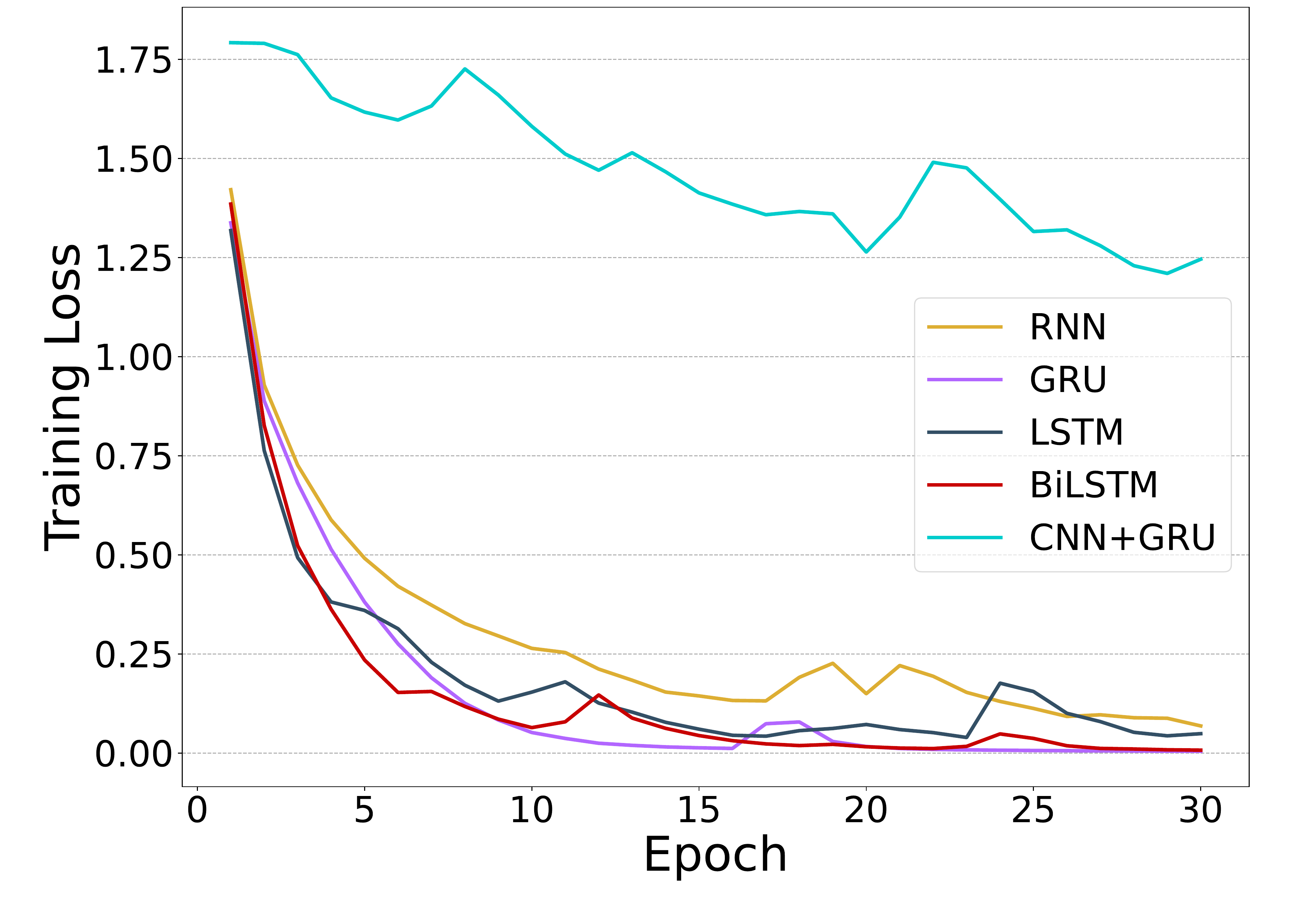}\label{subfig:NTU-HAR-GP2}}
	\subfigure[NTU-Fi Human-ID]{
    	\includegraphics[width=0.235\textwidth, angle=0]{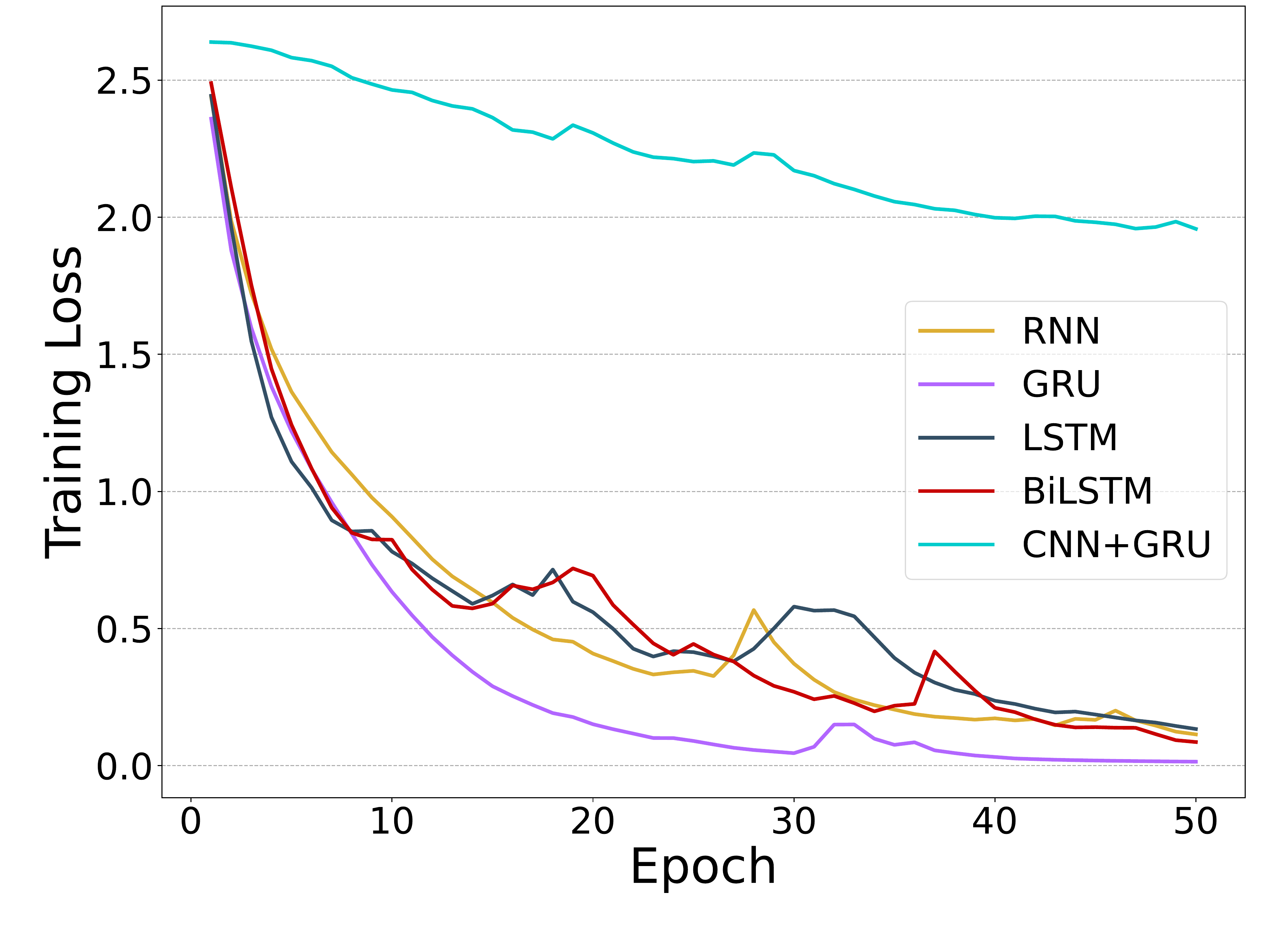}}
	\caption{\label{fig:training-loss-rnn}The training losses of RNN-based models for the four datasets.}
\end{figure*}

			
        
\subsection{Analysis}\label{sec:exp-analytics}
\textbf{Convergence of Deep Models.} Though all models converge eventually, their training difficulties are different and further affect their practical usage. To compare their convergence difficulties, we show the training losses of MLP, CNN-5, ViT, and RNN in terms of epochs in Figure~\ref{fig:training-loss}. It is noted that CNN converges very fast within 25 epochs for four datasets, and MLP also converges at a fast speed. The transformer requires more epochs of training since it consists of more model parameters. In comparison, RNN hardly converges on UT-HAR and Widar, and converges slower on NTU-Fi. Then we further explore the convergence of RNN-based models, including GRU, LSTM, BiLSTM, and CNN+GRU in Figure~\ref{fig:training-loss-rnn}. Though there show strong fluctuations during the training phase of GRU, LSTM, and BiLSTM, these three models can achieve much lower training loss. Especially, GRU achieves the lowest loss among all RNN-based methods. For CNN+GRU, the training phase is more stable but its convergence loss is larger than others. 

\textbf{How Transfer Learning Matters.} We further draw the training losses of all models on NTU-Fi Human-ID with pre-trained parameters of NTU-Fi HAR in Figure~\ref{fig:transfer-loss}. Compared to the training procedures of randomly-initialized models in Figures ~\ref{subfig:NTU-HAR-GP1} and~\ref{subfig:NTU-HAR-GP2}, the convergence can be achieved and even become much more stable. We can draw two conclusions from these results: (a) the feature extractors of these models are transferable across two similar tasks; (b) the fluctuations of training losses are caused by the feature extractor since only the classifier is trained for the transfer learning settings. 

\begin{figure}[t]
	\centering
    \subfigure[Accuracy]{
    	\includegraphics[width=0.232\textwidth, angle=0]{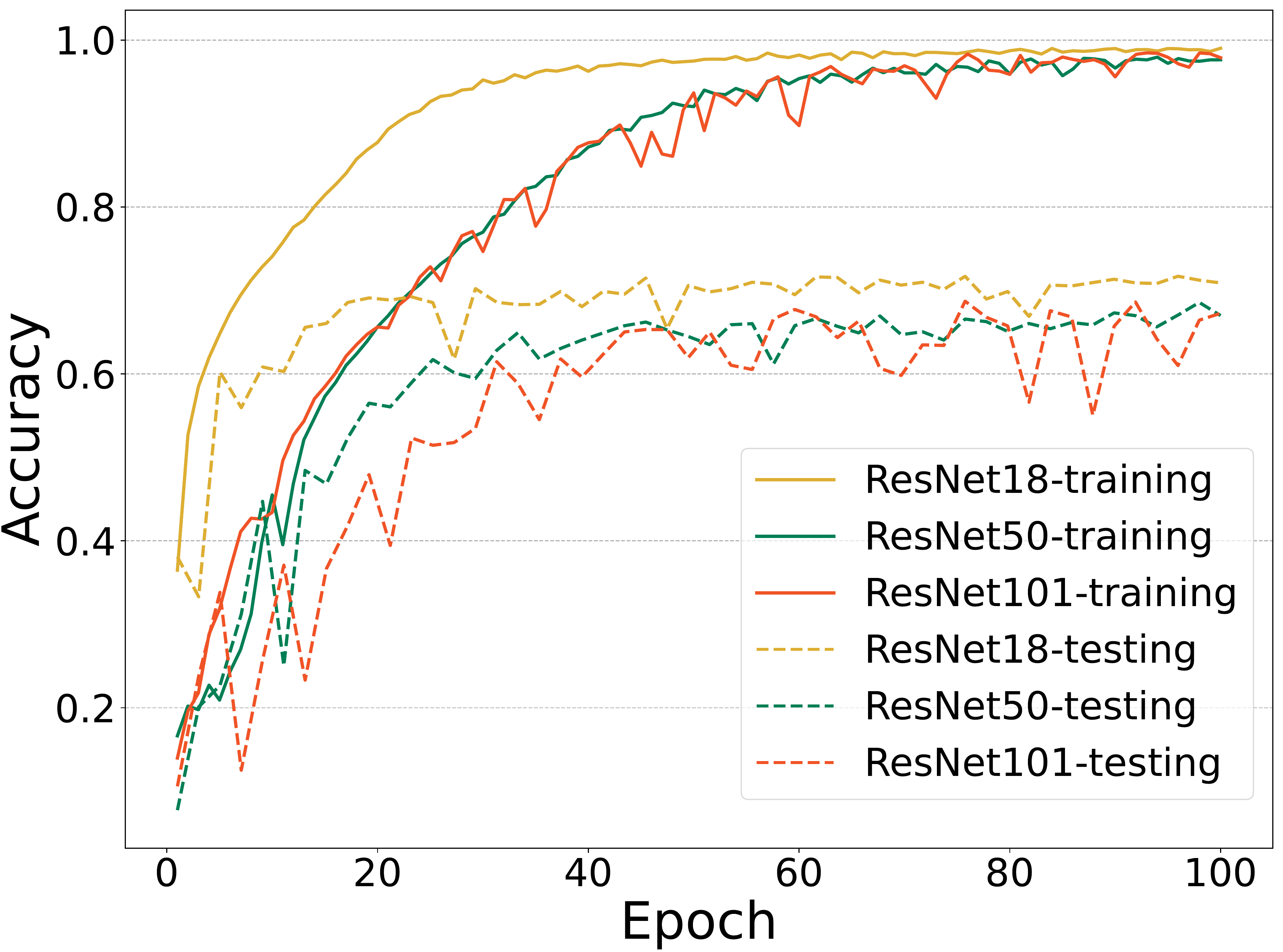}\label{subfig:widar-resnet-acc}}
    \subfigure[Accuracy]{
    	\includegraphics[width=0.232\textwidth, angle=0]{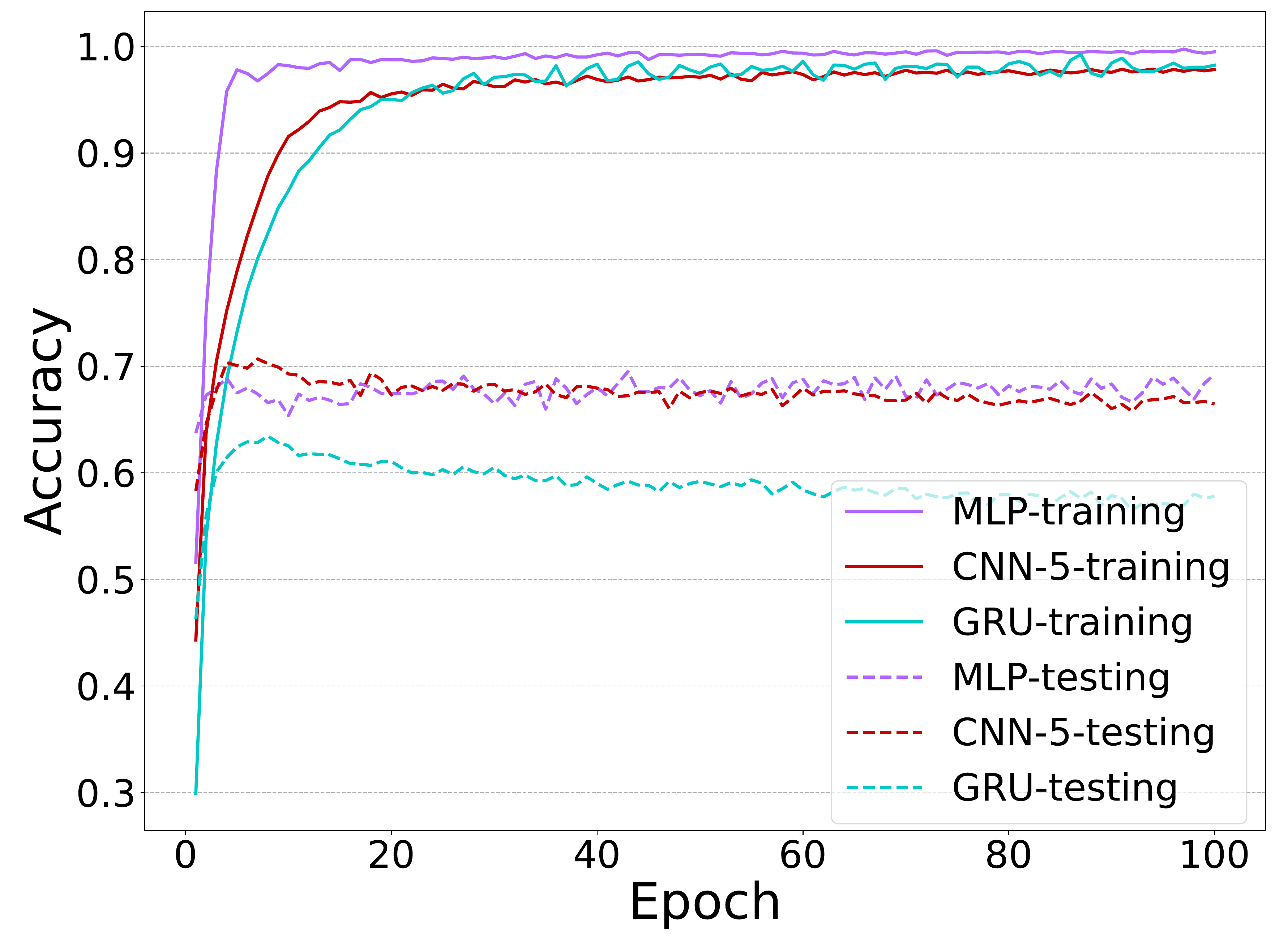}\label{subfig:widar-other-net-acc}}
	\caption{\label{fig:widar-acc}The training procedures of deep learning models on Widar data in terms of training and testing accuracy.}
\end{figure}
\textbf{Poor Performance of Deep CNN on Widar.}
In Table~\ref{tab:overall}, a noticeable phenomenon is that the ResNet-18/50/101 cannot generalize well on Widar data, only achieving 17.91\%, 19.47\%, and 14.47\%, respectively. In visual recognition, a deeper network should perform better on large-scale datasets ~\cite{he2016deep}. Then we have the question: is the degeneration of these deep models caused by underfitting or overfitting in WiFi sensing? We seek the reason by plotting their training losses in Figure~\ref{fig:widar-acc}. Figure~\ref{subfig:widar-resnet-acc} shows that even though the training accuracy has been almost 100\%, the testing accuracy remains low, under 20\%. Whereas, other networks (MLP, CNN, GRU) have similar training accuracy while the testing accuracy is increased to over 60\%. This indicates that the degrading performances of ResNets are caused by overfitting, and different domains in Widar~\cite{zhang2021widar3} might be the main reasons. This discovery tells us that very deep networks are prone to suffer from overfitting for cross-domain tasks and may not be a good choice for current WiFi sensing applications due to their performance and computational overhead.

\begin{figure}[t]
	\centering
	\subfigure[Training using Adam]{
    	\includegraphics[width=0.232\textwidth, angle=0]{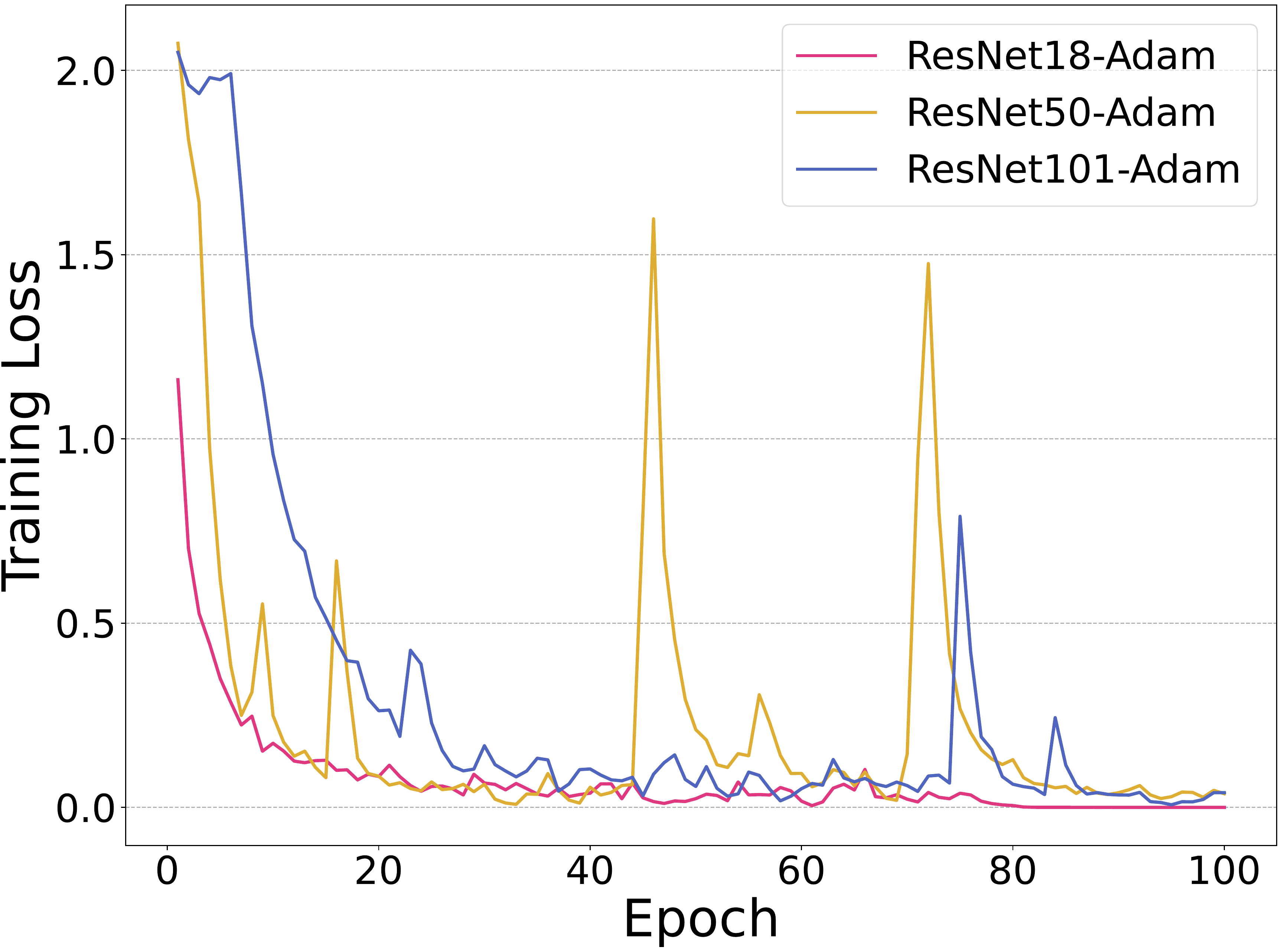}\label{subfig:resnet-adam}}
    \subfigure[Training using SGD]{
    	\includegraphics[width=0.232\textwidth, angle=0]{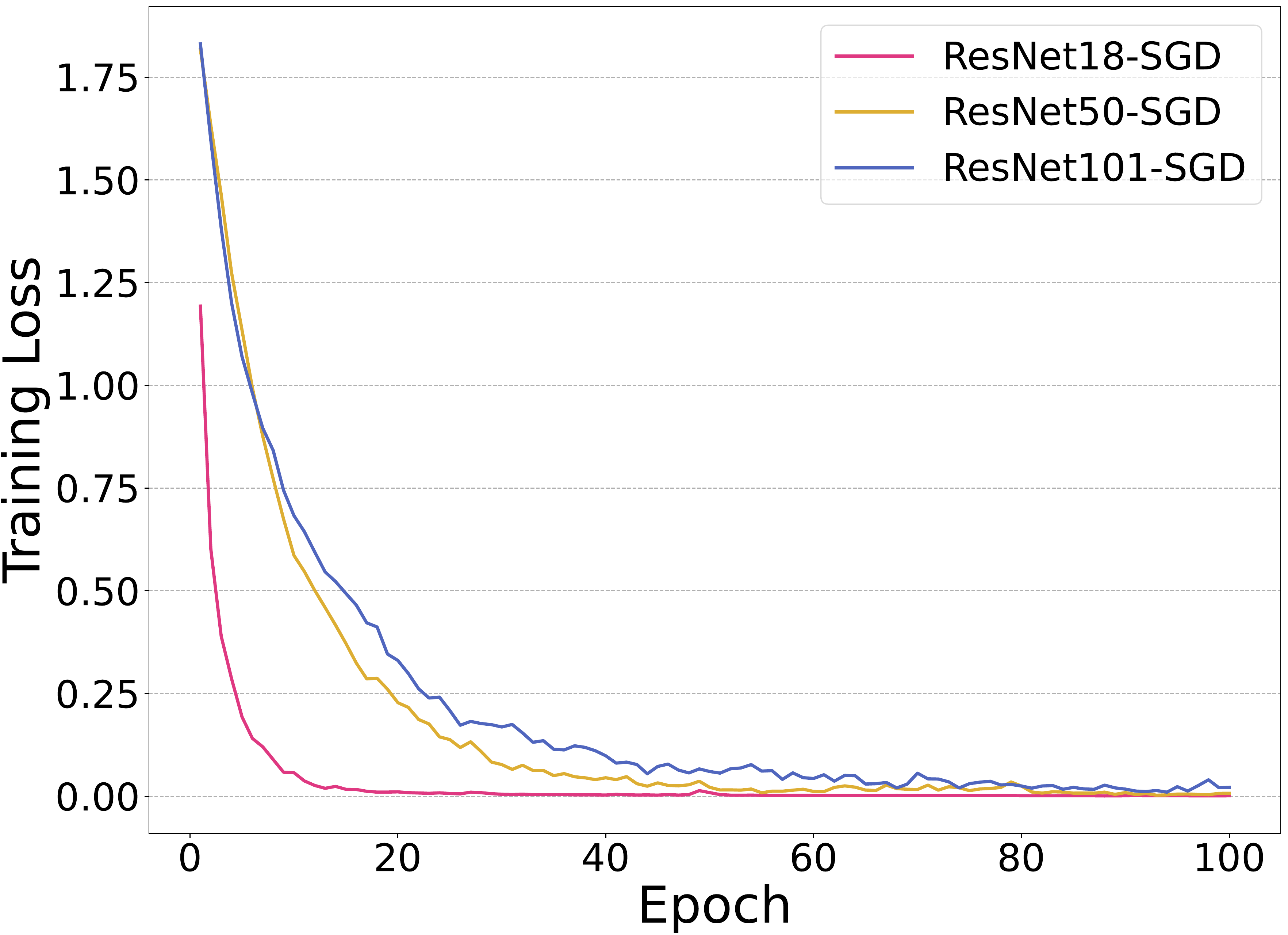}\label{subfig:resnet-sgd}}
	\caption{\label{fig:optimizer}The training procedures of ResNet-18/50/101 using Adam and SGD optimizers on UT-HAR.}
\end{figure}

\textbf{Choices of Optimizer.}
During the training phase, we find that though Adam can help models converge at a fast speed, it also leads to much training instability, especially for the very deep neural networks. In Figure~\ref{subfig:resnet-adam}, we can see that ResNet-18 converges stably but ResNet-50 and ResNet-101 have fluctuating losses every 20-30 epochs.
This might be caused by the dramatically changing values of WiFi data and its adaptive learning rate of Adam~\cite{kingma2014adam}. Then we consider changing the optimizer from Adam to a more stable optimizer, Stochastic Gradient Descent (SGD). In Figure~\ref{subfig:resnet-sgd}, we find that the training procedure becomes more stable. This implies that if a very deep model is utilized in WiFi sensing, the SGD should be a better choice. If a simple model is sufficient for the sensing task, then Adam can enforce the model to converge better and faster.

\section{Discussions and Summary}\label{sec:summary}
Having analyzed the empirical results and the characteristics of deep learning models for WiFi sensing, we summarize the experiences and observations that facilitate future research on model design, model training, and real world use case:
\begin{itemize}
    \item \textbf{Model Choices.} We recommend CNN, GRU, and BiLSTM due to their high performance, low computational cost, and small parameter size. The shallow models have achieved remarkable results for activity recognition, gesture recognition, and human identification, while the very deep models confront the overfitting issue, especially for cross-domain scenarios.
    \item \textbf{Optimization.} We recommend using Adam or SGD optimizer. The Adam optimizer enforces the model to converge at a fast speed but sometimes it causes instability of training. When such a situation happens, the SGD is a more secure way but the hyper-parameters of SGD (\textit{i.e.}, the learning rate and momentum) need to be manually specified and tuned.
    \item \textbf{Advice on Transfer Learning Applications.} We recommend applying transfer learning when the task is similar to existing applications and the same CSI sensing platform is employed. The pre-trained parameters provide a good initialization and better generalization ability. CNN, MLP, and BiLSTM have superior transferability.
    \item \textbf{Advice on Unsupervised Learning.} We recommend applying unsupervised learning to initialize the model for similar tasks since unsupervised learning extracts more generalizable features than transfer learning. CNN, MLP, and ViT are more suitable in the unsupervised learning framework in general.
\end{itemize}

\section{Grand Challenges and Future Directions}
Deep learning still keeps booming in many research fields and continuously empowers more challenging applications and scenarios. Based on the new progress, we look into the future directions of deep learning for WiFi sensing and summarize them as follows.

\textbf{Data-efficient learning.}
As CSI data is expensive to collect, data-efficient learning methods should be further explored. Existing works have utilized few-shot learning, transfer learning, and domain adaptation, which yield satisfactory results in a new environment with limited training samples. However, since the testing scenarios are simple, the transferability of these models cannot be well evaluated. In the future, meta-learning and zero-shot learning can further help learn robust features across environments and tasks.

\textbf{Model compression or lightweight model design.}
In the future, WiFi sensing requires real-time processing for certain applications, such as vital sign monitoring~\cite{hu2022resfi}. To this end, model compression techniques can play a crucial role, such as model pruning~\cite{chen2019cooperative}, quantization~\cite{chen2019metaquant} and distillation~\cite{yang2020mobileda}, which decreases the model size via an extra learning step. The lightweight model design is also favorable, such as the EfficientNet~\cite{tan2019efficientnet} in computer vision that is designed from scratch by balancing network depth, width, and resolution.

\textbf{Multi-modal learning.}
WiFi sensing is ubiquitous, cost-effective, and privacy-preserving, and can work without the effect of illumination and part of occlusion, which is complementary to the existing visual sensing technique. To achieve robust sensing 24/7, multiple modalities of sensing data should be fused using multi-modal learning. WiVi~\cite{zou2019wificv} pioneers human activity recognition by integrating WiFi sensing and visual recognition. Multi-modal learning can learn joint features from multiple modalities and make decisions by choosing reliable modalities.

\textbf{Cross-modal learning.}
WiFi CSI data describes the surrounding environment that can also be captured by cameras. Cross-modal learning aims to supervise or reconstruct one modality from another modality, which helps WiFi truly ``see'' the environment and visualize them in videos. Wi2Vi~\cite{kefayati2020wi2vi} manages to generate video frames by CSI data and firstly achieves cross-modal learning in WiFi sensing. The human pose is then estimated by supervising the model by the pose landmarks of OpenPose~\cite{wang2019person}. In the future, cross-modal learning may enable the WiFi model to learn from more supervisions such as radar and Lidar.

\textbf{Model robustness and security for trustworthy sensing.}
When deploying WiFi sensing models in the real world, the model should be secure to use. Adversarial attacks have raised attentions in video-based human sensing~\cite{liu2020adversarial}. Nevertheless, existing works study the accuracy of models but few pay attention to the security issue. First, during the communication, the sensing data may leak the privacy of users. Second, if any adversarial attack is made on the CSI data, the modal can perform wrongly and trigger the wrong actions of smart appliances. RobustSense seeks to overcome adversarial attacks by augmentation and adversarial training~\cite{yang2022autofi}. EfficientFi proposes a variational auto-encoder to quantize the CSI for efficient and robust communication. WiFi-ADG~\cite{zhou2019adversarial} protects the user privacy by enforcing the data not recognizable by general classifiers. More works should be focused on secure WiFi sensing and establish trustworthy models for large-scale sensing, such as federated learning.

\textbf{Complicated human activities and behaviors analytics.}
While current methods have shown prominent recognition accuracy for single activities or gestures, human behavior is depicted by more complicated activities. For example, to indicate if a patient may have a risk of Alzheimer’s disease, the model should record the routine and analyze the anomaly activity, which is still difficult for existing approaches. Precise user behavior analysis can contribute to daily healthcare monitoring and behavioral economics. 

\textbf{Model interpretability for a physical explanation.}
Model-based and learning-based methods develop fast but in a different ways. Recent research has investigated the interpretability of deep learning models that looks for the justifications of classifiers. In WiFi sensing, if the model is interpreted well, there may exist a connection between the data-driven model and the physical model. The modal interpretability may inspire us to develop new theories of physical models for WiFi sensing, and oppositely, the existing model (\textit{e.g.}, Fresnel Zone) may enable us to propose new learning methods based on the physical models. It is hoped that two directions of methods can be unified theoretically and practically.
\section{Conclusion}\label{sec:conclusion}
Deep learning methods have been proven to be effective for challenging applications in WiFi sensing, yet these models exhibit different characteristics on WiFi sensing tasks and a comprehensive benchmark is highly demanded. To this end, this work reviews the recent progress on deep learning for WiFi human sensing, and benchmarks prevailing deep neural networks and deep learning strategies on WiFi CSI data across different platforms. We summarize the conclusions drawn from the experimental observations, which provide valuable experiences for model design in practical WiFi sensing applications. Last but not least, the grand challenges and future directions are proposed to imagine the research issues emerging from future large-scale WiFi sensing scenarios.



\bibliographystyle{IEEEtran}
\bibliography{egbib}



%


\end{document}